\documentclass{article}
\usepackage{PRIMEarxiv}
\usepackage[utf8]{inputenc}
\usepackage[T1]{fontenc}
\usepackage{hyperref}
\usepackage{url}
\usepackage{booktabs}
\usepackage{amsfonts}
\usepackage{amsmath,amssymb}
\usepackage{nicefrac}
\usepackage{microtype}
\usepackage{graphicx}
\usepackage{enumitem}
\usepackage{array}
\usepackage{xcolor}
\usepackage[section]{placeins}
\usepackage{float}
\usepackage{needspace}
\usepackage{longtable}
\usepackage{multirow}
\usepackage{pdflscape}
\hypersetup{hidelinks}
\usepackage{caption}
\graphicspath{{figures/}}
\title{Uncheatable Eval: Dynamic Compression-Based Evaluation of Language Models}
\author{Kaifeng Tan\textsuperscript{1}\hspace{1.2cm}
Yudong Li\textsuperscript{2}\hspace{1.2cm}
Linlin Shen\textsuperscript{1} \\[0.7em]
\normalfont\small
\textsuperscript{1}\,Shenzhen University\hspace{1.2cm}
\textsuperscript{2}\,Tsinghua University}
\begin{document}
\maketitle

\begin{abstract}
Modern large language models are pretrained on massive datasets, making it difficult to prevent benchmark data from entering their training sets and undermining the reliability of evaluation results. Reliable evaluation is particularly challenging for base models, whose limited instruction-following ability complicates task-based assessment. We introduce \emph{Uncheatable Eval}, a dynamic benchmark that regularly collects newly published text to evaluate base language models and reduce the risk of data contamination. Drawing on the relationship between a model's predictive ability and its ability to compress data losslessly, we use compression rate to evaluate how well models predict new text. We evaluate 80 models across 14 text categories, study how compression changes with context length, and examine the correlation between compression rate and zero-shot MMLU accuracy. Our results yield three main findings: (1) compression performance follows a consistent scaling trend with model size; (2) attention-based, hybrid, and recurrent models differ in how their compression performance changes as more context becomes available; and (3) lower compression rates are strongly associated with higher zero-shot MMLU accuracy. Code is available at \url{https://github.com/Jellyfish042/uncheatable_eval}.
\end{abstract}

\keywords{language model evaluation \and data contamination \and dynamic benchmarks \and data compression}

\begin{figure}[H]
\centering
\includegraphics[width=\linewidth]{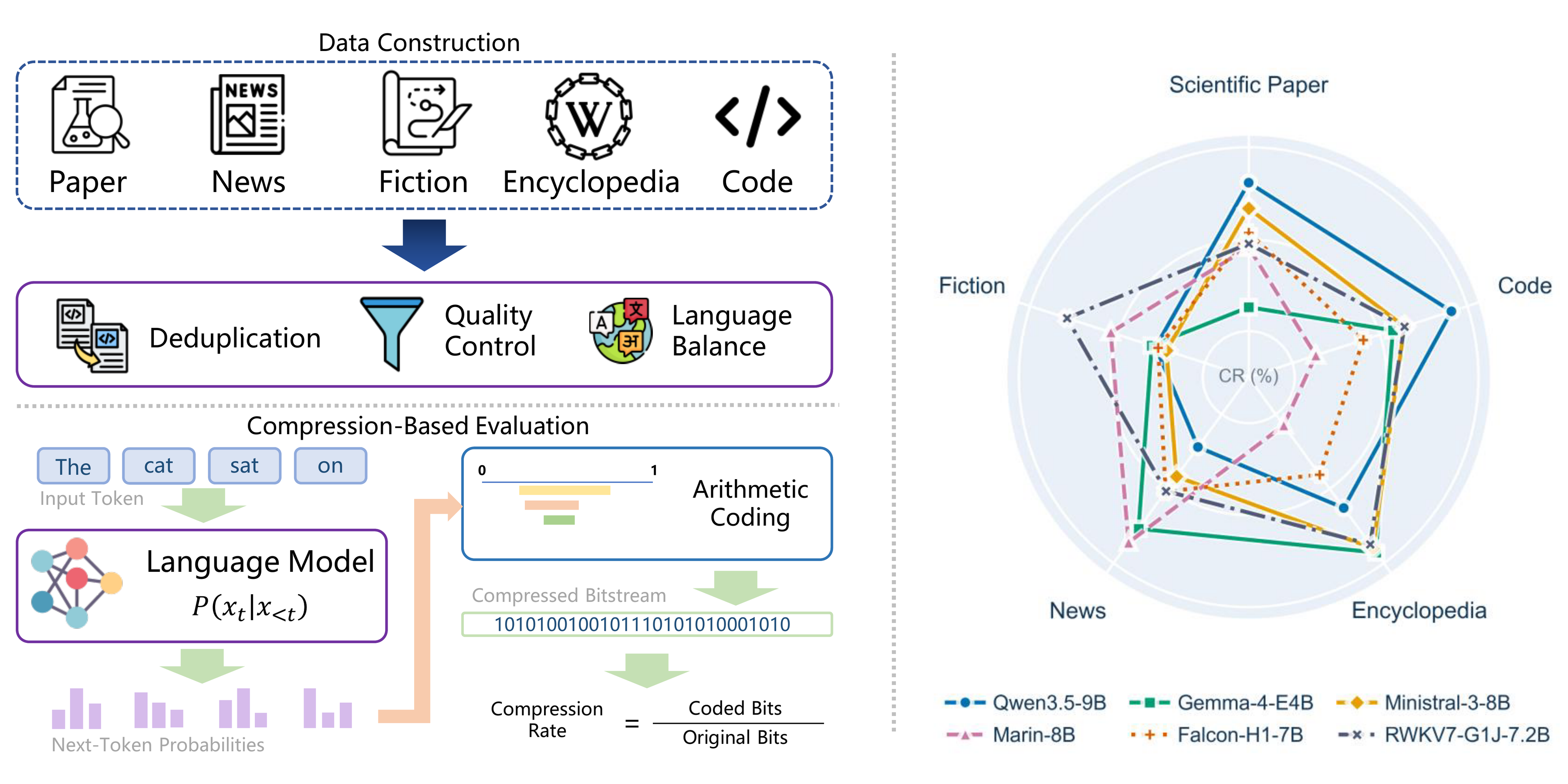}
\caption{Overview of Uncheatable Eval. Left: Data construction and compression-rate evaluation. Right: Compression rates of six representative models across five text types.}
\label{fig:overview}
\end{figure}

\section{Introduction}
\label{sec:introduction}

Effective evaluation is essential to the development of large language models. However, the vast internet corpora used to train these models may contain benchmark questions, answers, and solutions. Such contamination can inflate scores through memorization, making it difficult to distinguish genuine generalization from recall. Longitudinal studies using model training cutoffs find evidence of this effect in code and mathematics data \cite{roberts2023time}; surveys and controlled studies show that contamination occurs at several semantic levels and is difficult to detect reliably \cite{xu2024benchmark,deng2024investigating,fu2025contamination}.

Dynamic benchmarks reduce this risk by collecting newly published data or generating new test examples. LiveBench and LiveCodeBench use recently released questions and code problems \cite{white2025livebench,jain2024livecodebench}. Other recent benchmarks continuously update their evaluation data to reflect changes in facts and knowledge \cite{xu2026livefact,kim2026oaks,liu2026rag}. Newer test data reduce contamination risk, while task-based evaluation can obscure base-model capabilities because base models have limited instruction-following ability.

We introduce Uncheatable Eval, a dynamic benchmark that measures how well pretrained language models predict newly published text (Figure~\ref{fig:overview}). Drawing on the relationship between prediction and lossless compression, we view each language model as a lossless compressor: its next-token probabilities can guide arithmetic coding, assigning shorter codes to more probable continuations \cite{deletang2023compression}. A model that predicts the observed text more accurately therefore requires fewer bits to encode it. We evaluate this ability through compression rate, calculated from the ideal code length relative to the original text size. This directly evaluates base models on new text without task instructions or reference answers.

To apply this approach across diverse forms of text, we collect newly published fiction, source code, scientific papers, news, and encyclopedia articles within specified time windows. We then process these texts through a data curation pipeline that combines quality filtering, deduplication, and text normalization, with balanced sampling across languages for the multilingual subset. Results for 80 models on the resulting datasets reveal a systematic relationship between model size and compression rate, well described by a power law with an additive constant. A 54-model analysis on longer scientific documents shows different changes in CR across attention-based, hybrid, and recurrent models as more preceding text becomes available. A comparison with zero-shot MMLU further reveals a strong association between better compression of technical text and higher MMLU accuracy.

Our main contributions can be summarized as follows:
\begin{enumerate}[leftmargin=*]
\item We introduce Uncheatable Eval, a dynamic benchmark that evaluates base language models through compression of newly published text collected from diverse sources and curated using a data processing pipeline developed for the benchmark.
\item We report 80 models spanning attention-based, hybrid, and recurrent families across 14 text categories, providing a broad comparison of compression performance across model sizes and text domains.
\item We analyze how CR varies with model size and document position and how it correlates with zero-shot MMLU accuracy.
\end{enumerate}

\section{Related Work}
\label{sec:related}

\paragraph{Data contamination.}
Benchmark questions and answers can enter model training data and inflate test scores \cite{deng2024investigating,li2024opensourcecontamination,sainz2023nlp}. Contamination ranges from exact copies to semantic overlap, so lexical matching alone is insufficient \cite{xu2024benchmark,chen2025benchmarking}. Comparisons across training cutoffs and controlled leakage experiments reveal score effects and detection limits \cite{roberts2023time,hidayat2025simulating}. Yet overlap does not have the same performance effect for every model, and inflated scores need not reorder leaderboards \cite{singh2024contamination,xiao2026leaderboards}. Other approaches detect contamination from model behavior \cite{dong2024generalization,fu2025contamination}, score multiple contamination levels \cite{xu2025dcr}, or control screening errors \cite{zhang2026controllable}. Proposed defenses vary test instances or interactions \cite{qian2024varbench,yu2024kieval,zhu2024inference,chai2026benchmarks}, limit benchmark exposure \cite{jacovi2023stop}, or design contamination-resistant datasets \cite{allawati2026contamination}. These studies motivate collecting new evaluation data and show that publication recency alone does not establish a clean test set.

\paragraph{Dynamic benchmarks.}
Dynamic benchmarks regularly add new test data to reduce the risk of contamination. LiveBench and LiveCodeBench collect recently released questions and programming problems \cite{white2025livebench,jain2024livecodebench}, while AntiLeakBench constructs questions using new real-world knowledge \cite{wu2025antileakbench}. Related benchmarks test whether models can use updated evidence, track changing facts, and adapt to new knowledge \cite{xu2026livefact,kim2026oaks,liu2026rag}.

FreshBench is the closest prior work to ours. It measures the probabilities models assign to text published at different times and their accuracy in predicting future events \cite{zhu2025freshbench}. Its main question is how performance varies with the age of the data relative to a model's release, including whether models perform worse on more recent data. Uncheatable Eval also evaluates newly published text, but focuses on comparing base models through compression rate. We also examine compression by model size and document position.


\paragraph{Compression-based evaluation.}
Language models can be evaluated directly by measuring how well they predict text. Paloma measures perplexity across diverse domains under consistent evaluation conditions \cite{magnusson2024paloma}. Del\'etang et al. explain the connection between prediction and lossless compression: a model that assigns higher probabilities to the observed text can encode it using fewer bits \cite{deletang2023compression}. Li et al. use compression to evaluate models on text published after their training cutoffs and examine how performance changes across time periods \cite{li2024compression}. Huang et al. find that models that compress text better also tend to score higher on knowledge, code, and mathematics benchmarks \cite{huang2024compression}. Uncheatable Eval extends this approach to a regularly updated benchmark covering diverse text types and model families.

\section{Uncheatable Eval}
\label{sec:goals}

\subsection{Data Collection and Curation}
\label{sec:data}

Uncheatable Eval collects newly published text for each evaluation to reduce the risk that models have already encountered the test data during training. The collection covers fiction, news, encyclopedia articles, scientific papers, source code, and documentation from publicly available sources. Table~\ref{tab:data-categories} lists the evaluation categories; \emph{Other} groups source-code files outside the named language categories.

The collected text is processed through a curation pipeline to improve quality and limit repetition. Quality filters remove samples that are too short or contain anomalous character sequences, while near-duplicate removal uses MinHash locality-sensitive hashing to limit repeated content. Unicode text is standardized using NFC normalization. For the non-English encyclopedia subset, sampling is balanced across languages to prevent the results from being dominated by a single language.

Because tokenizers can assign different numbers of tokens to the same text, a length limit based on one tokenizer may leave samples too long for another. We therefore check each sample with a diverse set of tokenizers and truncate it to meet a common token budget under all of them. This produces a shared text sample for evaluation while accounting for differences in tokenization.
\begin{table}[htbp]
\centering
\caption{Evaluation categories and labels used in the result tables.}
\label{tab:data-categories}
\small
\setlength{\tabcolsep}{6pt}
\renewcommand{\arraystretch}{1.10}
\begin{tabular}{@{}>{\raggedright\arraybackslash}p{0.31\linewidth}>{\raggedright\arraybackslash}p{0.31\linewidth}>{\raggedright\arraybackslash}p{0.31\linewidth}@{}}
\toprule
\textbf{General text} & \textbf{Scientific Paper} & \textbf{Code} \\
\midrule
English fiction (EF) & Biology preprints (Bio) & C++ \\
News & Computer science papers (CS) & JavaScript (JS) \\
English encyclopedia (EE) & Mathematics papers (Math) & Markdown (MD) \\
Non-English encyclopedia (NEE) & Other scientific papers (Sci-O) & Other \\
& Physics papers (Phys) & Python (Py) \\
\bottomrule
\end{tabular}
\end{table}

\subsection{Compression-Based Evaluation}
\label{sec:metric}

Uncheatable Eval measures how well a model predicts text by treating the model as a lossless compressor. In arithmetic coding, the model's next-token probabilities determine how the text is encoded: more probable sequences require fewer bits \cite{deletang2023compression}. A model that predicts the observed text more accurately can therefore compress it more efficiently. We measure this efficiency using the ideal code length implied by the model's probabilities.

Specifically, let $s$ be a document and $x_{1:T}$ its token sequence under the evaluated model's tokenizer. The model predicts each token from the preceding tokens, assigning the document the probability
\begin{equation}
p_\theta(x_{1:T})=\prod_{t=1}^{T}p_\theta(x_t\mid x_{<t}).
\label{eq:sequence-probability}
\end{equation}
The corresponding ideal code length, in bits, is
\begin{equation}
B_\theta(s)=-\log_2 p_\theta(x_{1:T})
=-\sum_{t=1}^{T}\log_2 p_\theta(x_t\mid x_{<t}).
\label{eq:bits}
\end{equation}
Thus, code length is the next-token negative log-likelihood expressed in base 2, linking the evaluation directly to the pretraining objective. It can be computed from the probabilities of the observed tokens without asking the model to follow instructions or generate an answer. We use this theoretical code length throughout; it excludes finite-precision coding overhead and the storage cost of model parameters.

To compare documents of different lengths and models with different tokenizers, we normalize the code length by the original document's UTF-8 size. If $n_{\mathrm{byte}}(s)$ is the number of UTF-8 bytes in $s$, its uncompressed size is $8n_{\mathrm{byte}}(s)$ bits. We define the compression rate (CR) as the percentage of this size needed for the ideal encoding:
\begin{equation}
\mathrm{CR}_\theta(s)=100\cdot\frac{B_\theta(s)}{8n_{\mathrm{byte}}(s)},
\label{eq:metrics}
\end{equation}
For example, a CR of 10\% means that the ideal encoding occupies one tenth of the original UTF-8 size. Lower CR indicates better compression and, on the same text, better prediction. The UTF-8 denominator puts models on a common scale; each score reflects the model--tokenizer pair.

For a collection of documents $\mathcal{S}$, compression rate is the ratio of their total ideal code length to their total original size:
\begin{equation}
\mathrm{CR}_\theta(\mathcal{S})=
100\cdot\frac{\sum_{s\in\mathcal{S}}B_\theta(s)}
{8\sum_{s\in\mathcal{S}}n_{\mathrm{byte}}(s)}.
\label{eq:aggregate}
\end{equation}
We use this byte-weighted rate to summarize performance within each category. To compare models across categories, we report an Overall score that takes the unweighted arithmetic mean of the category rates. This gives each category equal weight, so categories containing more bytes do not dominate the overall comparison.

\subsection{Long-Context Evaluation}
\label{sec:implementation}

Long-context evaluation examines how compression changes as a model processes more of a document. We retain the loss assigned to each token and align it to the token's UTF-8 bytes. Let $\ell_{s,t}=-\ln p_\theta(x_{s,t}\mid x_{s,<t})$ be the token loss for document $s$, and let $m_{s,t}$ be the number of UTF-8 bytes produced by that token. We assign $\ell_{s,t}/m_{s,t}$ to each of its bytes. The byte-wise compression rate at position $i$ is then
\begin{equation}
\mathrm{CR}_{\theta,i}=\frac{100}{8\ln 2}\cdot\frac{1}{N_i}
\sum_{s:\,i\leq |s|}\frac{\ell_{s,t_s(i)}}{m_{s,t_s(i)}},
\label{eq:byte-wise-cr}
\end{equation}
where $t_s(i)$ is the token containing byte $i$ and $N_i$ is the number of documents that reach that position. This construction compares the same byte positions across models with different tokenizers.

\section{Experiments and Results}
\label{sec:experiments}

\subsection{Experimental setup}

\paragraph{Data and models.}
We evaluate 80 models on a July 2026 dataset with 14 text categories collected from publicly available sources. Each category contains 500 samples, for a total of 7,000 samples. During data preparation, each sample is truncated to at most 3,584 tokens under each of eight tokenizers, producing a shared text sample for all evaluated models. Table~\ref{tab:main-results} presents 25 popular models, and Appendix~\ref{app:complete-results} reports all 80.

\paragraph{Evaluation metrics.}
We report compression rate (CR) as a percentage, with lower values indicating better compression. Within each category, CR is computed from the total ideal code length and total UTF-8 size using Eq.~\ref{eq:aggregate}. For the scaling analyses, we instead combine code lengths and byte counts across categories before computing CR. We also analyze five text types separately: Scientific Paper, Code, Encyclopedia, News, and English Fiction. These contain five, five, two, one, and one categories, respectively.

\paragraph{Long-context evaluation.}
The main evaluation truncates each July 2026 document to at most 3,584 tokens under each tokenizer. For long-context evaluation, we retain up to 32,768 UTF-8 bytes of the same documents. We evaluate 54 models on 500 documents from each of four scientific-paper categories: computer science, mathematics, physics, and other scientific papers. All four categories contain documents reaching 32~KiB. We compute byte-wise CR within each category using Eq.~\ref{eq:byte-wise-cr}; pooled curves weight each category by its number of contributing documents at each byte position.

\Needspace{430pt}
\subsection{Main results}

\Needspace{410pt}
\begingroup
\fontsize{7}{8.6}\selectfont
\setlength{\tabcolsep}{1.5pt}
\renewcommand{\arraystretch}{1.04}
\begin{longtable}{@{}>{\raggedright\arraybackslash}p{3.4cm}>{\raggedleft\arraybackslash}p{0.95cm}*{14}{>{\raggedleft\arraybackslash}p{0.64cm}}>{\raggedleft\arraybackslash}p{0.95cm}@{}}
\caption{Main results of Uncheatable Eval. All score cells are compression rate (CR, \%); lower is better.}\label{tab:main-results}\\
\toprule
\multirow{2}{*}{\raggedright Model} & \multirow{2}{*}{\raggedleft Params.} & \multicolumn{4}{c}{General text} & \multicolumn{5}{c}{Scientific Paper} & \multicolumn{5}{c}{Code} & Overall \\
\cmidrule(lr){3-6}\cmidrule(lr){7-11}\cmidrule(lr){12-16}\cmidrule(lr){17-17}
& & EF & News & EE & NEE & Bio & CS & Math & Sci-O & Phys & C++ & JS & MD & Other & Py & Overall \\
\midrule
\endfirsthead
\caption[]{Main results of Uncheatable Eval. All score cells are compression rate (CR, \%); lower is better. (continued)}\\
\toprule
\multirow{2}{*}{\raggedright Model} & \multirow{2}{*}{\raggedleft Params.} & \multicolumn{4}{c}{General text} & \multicolumn{5}{c}{Scientific Paper} & \multicolumn{5}{c}{Code} & Overall \\
\cmidrule(lr){3-6}\cmidrule(lr){7-11}\cmidrule(lr){12-16}\cmidrule(lr){17-17}
& & EF & News & EE & NEE & Bio & CS & Math & Sci-O & Phys & C++ & JS & MD & Other & Py & Overall \\
\midrule
\endhead
\midrule
\multicolumn{17}{r}{\emph{Continued on next page}} \\
\endfoot
\bottomrule
\endlastfoot
\midrule
\multicolumn{17}{@{}l}{\textbf{>20B}} \\
\addlinespace[1.5pt]
Gemma-4-31B & 31.3 & \textbf{9.17} & \textbf{7.46} & \textbf{6.98} & \textbf{6.09} & \textbf{6.43} & 6.59 & 5.98 & 6.16 & 6.31 & \textbf{3.16} & \textbf{3.70} & \textbf{7.86} & \textbf{3.98} & \textbf{4.15} & \textbf{6.00} \\
Qwen3.5-35B-A3B-Base & 34.7 & 9.73 & 8.25 & 7.61 & 7.49 & \textbf{6.43} & \textbf{6.52} & \textbf{5.78} & \textbf{6.09} & \textbf{6.24} & 3.33 & 3.75 & 7.92 & 4.04 & 4.24 & 6.24 \\
Gemma-4-26B-A4B & 25.8 & 9.52 & 7.68 & 7.22 & 6.41 & 6.62 & 6.92 & 6.22 & 6.44 & 6.56 & 3.29 & 3.87 & 8.34 & 4.17 & 4.38 & 6.26 \\
Mistral-Small-24B & 24.0 & 9.36 & 7.85 & 7.34 & 6.82 & 6.44 & 6.79 & 6.08 & 6.30 & 6.33 & 3.44 & 4.04 & 8.39 & 4.30 & 4.49 & 6.28 \\
Seed-OSS-36B-Base & 36.2 & 9.61 & 8.02 & 7.48 & 8.51 & 6.52 & 6.83 & 5.97 & 6.36 & 6.57 & 3.26 & 3.89 & 8.12 & 4.07 & 4.28 & 6.39 \\
Nemotron-3-Nano-30B-A3B & 31.6 & 9.70 & 7.96 & 7.33 & 7.40 & 6.60 & 6.84 & 6.19 & 6.43 & 6.49 & 3.44 & 4.15 & 8.49 & 4.44 & 4.50 & 6.43 \\
OLMo-3-32B & 32.2 & 9.60 & 8.05 & 7.71 & 9.24 & 6.58 & 6.87 & 6.18 & 6.42 & 6.60 & 4.09 & 4.85 & 8.88 & 5.07 & 4.91 & 6.79 \\
\midrule
\multicolumn{17}{@{}l}{\textbf{$\sim$14B}} \\
\addlinespace[1.5pt]
RWKV7-G1J-13.3B & 13.3 & \textbf{9.30} & \textbf{8.01} & \textbf{7.45} & \textbf{6.87} & 6.62 & \textbf{6.68} & 6.13 & \textbf{6.32} & 6.48 & 3.59 & \textbf{4.07} & \textbf{8.14} & \textbf{4.30} & \textbf{4.42} & \textbf{6.31} \\
Ministral-3-14B & 13.9 & 9.86 & 8.22 & 7.64 & 6.99 & \textbf{6.44} & 6.80 & \textbf{6.03} & 6.33 & \textbf{6.36} & \textbf{3.55} & 4.21 & 8.60 & 4.46 & 4.65 & 6.44 \\
\midrule
\multicolumn{17}{@{}l}{\textbf{$\sim$7B}} \\
\addlinespace[1.5pt]
Qwen3.5-9B-Base & 9.0 & 10.15 & 8.60 & 8.02 & 8.24 & 6.67 & \textbf{6.76} & \textbf{6.06} & \textbf{6.34} & \textbf{6.50} & 3.65 & \textbf{4.08} & \textbf{8.28} & \textbf{4.33} & \textbf{4.51} & \textbf{6.59} \\
Ministral-3-8B & 8.9 & 10.20 & 8.42 & 7.90 & \textbf{7.34} & \textbf{6.64} & 7.02 & 6.25 & 6.54 & 6.58 & 3.71 & 4.42 & 8.93 & 4.66 & 4.84 & 6.68 \\
RWKV7-G1J-7.2B & 7.2 & \textbf{9.69} & 8.34 & 7.84 & 7.48 & 6.94 & 7.05 & 6.54 & 6.67 & 6.84 & 3.91 & 4.44 & 8.75 & 4.67 & 4.81 & 6.71 \\
Meta-Llama-3.1-8B & 8.0 & 9.89 & \textbf{8.01} & \textbf{7.66} & 7.45 & 6.83 & 7.35 & 6.36 & 6.73 & 6.77 & 3.94 & 4.73 & 9.39 & 4.88 & 5.08 & 6.79 \\
Qwen3-8B-Base & 8.2 & 10.21 & 8.61 & 8.19 & 8.12 & 6.92 & 7.22 & 6.15 & 6.68 & 6.80 & \textbf{3.63} & 4.33 & 8.99 & 4.57 & 4.74 & 6.80 \\
Falcon-H1-7B-Base & 7.6 & 10.16 & 8.33 & 8.00 & 9.09 & 6.88 & 7.23 & 6.17 & 6.65 & 6.71 & 3.96 & 4.69 & 9.41 & 4.90 & 5.12 & 6.95 \\
marin-8b-base & 8.0 & 9.91 & 8.02 & 7.86 & 10.45 & 6.91 & 7.25 & 6.44 & 6.75 & 6.75 & 4.33 & 5.17 & 9.63 & 5.25 & 5.46 & 7.16 \\
\midrule
\multicolumn{17}{@{}l}{\textbf{$\sim$3B}} \\
\addlinespace[1.5pt]
Qwen3.5-4B-Base & 4.2 & 10.65 & 9.05 & 8.49 & 8.95 & 6.99 & \textbf{7.07} & \textbf{6.41} & \textbf{6.66} & \textbf{6.83} & \textbf{3.98} & \textbf{4.43} & \textbf{8.79} & \textbf{4.66} & \textbf{4.83} & \textbf{6.98} \\
RWKV7-G1J-2.9B & 2.9 & \textbf{10.02} & \textbf{8.67} & \textbf{8.22} & \textbf{8.05} & 7.28 & 7.44 & 6.96 & 7.05 & 7.23 & 4.25 & 4.86 & 9.36 & 5.05 & 5.21 & 7.12 \\
Ministral-3-3B & 3.8 & 10.89 & 8.90 & 8.41 & 8.10 & \textbf{6.98} & 7.44 & 6.69 & 6.94 & 7.00 & 4.08 & 4.87 & 9.62 & 5.08 & 5.25 & 7.16 \\
Falcon-H1-3B-Base & 3.1 & 10.84 & 8.87 & 8.65 & 10.25 & 7.44 & 7.72 & 6.64 & 7.12 & 7.22 & 4.52 & 5.27 & 10.27 & 5.53 & 5.68 & 7.57 \\
\midrule
\multicolumn{17}{@{}l}{\textbf{$<2$B}} \\
\addlinespace[1.5pt]
RWKV7-G1J-1.5B & 1.5 & \textbf{10.56} & \textbf{9.16} & \textbf{8.76} & \textbf{8.84} & 7.71 & 7.91 & 7.46 & 7.50 & 7.69 & 4.69 & 5.35 & 10.10 & 5.53 & 5.67 & \textbf{7.64} \\
Qwen3.5-2B-Base & 1.9 & 11.51 & 9.84 & 9.28 & 10.09 & \textbf{7.62} & \textbf{7.75} & \textbf{7.08} & \textbf{7.30} & \textbf{7.51} & \textbf{4.67} & \textbf{5.14} & \textbf{9.82} & \textbf{5.35} & \textbf{5.52} & 7.75 \\
Falcon-H1-1.5B-Deep-Base & 1.6 & 11.27 & 9.36 & 9.18 & 11.33 & 7.90 & 8.24 & 7.10 & 7.58 & 7.83 & 5.07 & 5.88 & 11.06 & 6.16 & 6.20 & 8.16 \\
Qwen3.5-0.8B-Base & 0.8 & 12.52 & 10.81 & 10.23 & 11.47 & 8.38 & 8.50 & 7.81 & 8.01 & 8.23 & 5.42 & 5.94 & 10.95 & 6.12 & 6.27 & 8.62 \\
Falcon-H1-0.5B-Base & 0.5 & 12.96 & 10.35 & 10.53 & 18.95 & 9.00 & 9.29 & 8.18 & 8.58 & 8.92 & 6.20 & 7.27 & 13.10 & 7.54 & 7.47 & 9.88 \\
\end{longtable}
\vspace{-0.5\baselineskip}
\noindent\parbox{\linewidth}{\footnotesize Abbreviations: EF, English fiction; EE/NEE, English/non-English encyclopedia; Bio, biology preprints; CS, computer science papers; Sci-O, other scientific papers; JS, JavaScript; MD, Markdown; Py, Python. Other denotes source code outside C++, JavaScript, and Python. Params. gives total parameters in billions. Overall is the unweighted mean of the 14 category CRs. Within each parameter-size group, bold marks the lowest displayed CR in each column, including ties.}
\endgroup

Table~\ref{tab:main-results} presents compression rates for 25 representative models across the 14 text categories, together with their Overall scores; Appendix~\ref{app:complete-results} reports the complete results for all 80 models. Qwen models \cite{yang2025qwen3,qwen2026qwen35} are consistently strong on scientific papers and code: Qwen3.5-35B-A3B-Base leads several scientific categories, while Qwen3.5 and Qwen3 remain among the strongest code compressors at their respective scales. RWKV7-G1J \cite{peng2025rwkv7} shows the complementary pattern. Its models perform particularly well on general text, including the non-English encyclopedia category, where the family often leads similarly sized alternatives.

Mixture-of-experts models also perform competitively with far fewer active parameters than total parameters. Qwen3.5-35B-A3B-Base, Gemma-4-26B-A4B \cite{gemmateam2026gemma4}, and Nemotron-3-Nano-30B-A3B \cite{nvidia2025nemotron3nano} reach compression rates comparable to dense models with similar total sizes, and Qwen3.5-35B-A3B-Base approaches the best Overall score in the table.

\FloatBarrier
\subsection{Long-context evaluation}

\begin{figure}[!htbp]
\centering
\includegraphics[width=\linewidth]{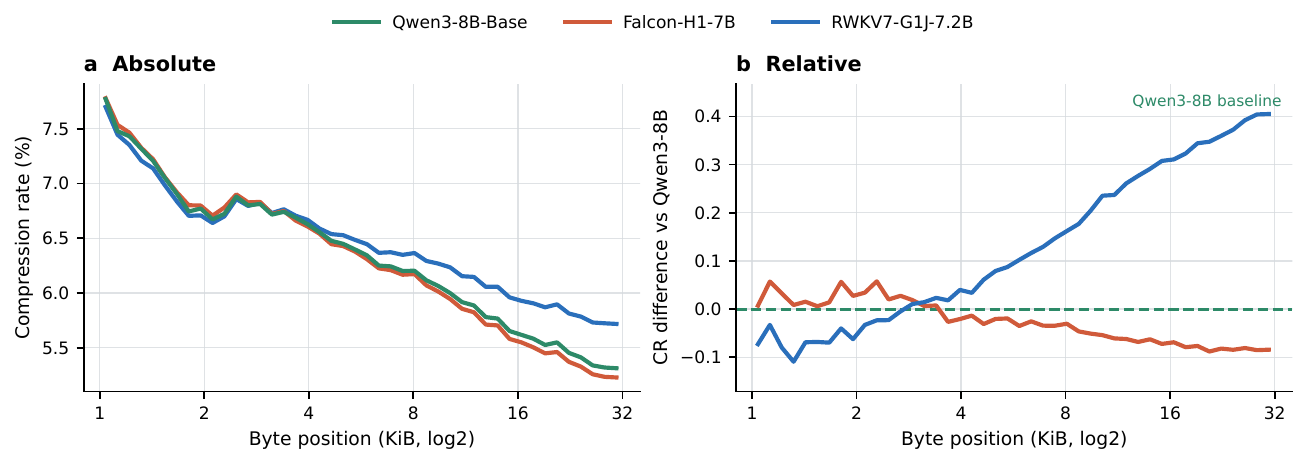}
\caption{Long-context compression for Qwen3-8B-Base, Falcon-H1-7B, and RWKV7-G1J-7.2B on four scientific-paper categories. (a) Byte-wise CR pooled over documents that reach each position, shown in logarithmically spaced bins from 1 to 32~KiB. (b) CR differences from Qwen3-8B-Base.}
\label{fig:long-context}
\end{figure}

We evaluate 54 models on documents from four scientific-paper categories and report the complete long-context results in Appendix~\ref{app:long-context-scale}. We focus here on Qwen3-8B-Base, Falcon-H1-7B, and RWKV7-G1J-7.2B as representatives of attention-based, hybrid, and recurrent architectures, respectively \cite{yang2025qwen3,zuo2025falconh1,peng2025rwkv7}. Figure~\ref{fig:long-context} compares the three models' byte-wise CR as document position increases from 1 to 32~KiB, both in absolute terms and relative to Qwen3-8B-Base.

\begin{table}[!htbp]
\centering
\small
\caption{Long-context slopes for similar-sized models near 7--8B on four scientific-document categories.}
\label{tab:long-context}
\setlength{\tabcolsep}{4.2pt}
\begin{tabular}{llrrrrrr}
\toprule
Architecture & Model & CS & Math & Physics & Other & Pooled & Gain \\
\midrule
Attention & Qwen3-8B-Base & -0.377 & -0.528 & -0.396 & -0.415 & -0.426 & 1.700 \\
Hybrid & Falcon-H1-7B & -0.404 & -0.562 & -0.424 & -0.440 & -0.454 & 1.807 \\
Recurrent & RWKV7-G1J-7.2B & -0.232 & -0.461 & -0.280 & -0.293 & -0.314 & 1.266 \\
\bottomrule
\end{tabular}
\end{table}

Figure~\ref{fig:long-context} shows that all three models achieve lower CR at later positions, but at different rates. To quantify this trend, we summarize the byte-wise rates in five non-overlapping intervals: 1--2, 2--4, 4--8, 8--16, and 16--32~KiB. Within each interval, we again weight every byte position by its document count. We then fit
\begin{equation}
\overline{\mathrm{CR}}_j=\alpha+\beta\log_2 x_j,
\label{eq:context-slope}
\end{equation}
where $\overline{\mathrm{CR}}_j$ is the weighted mean for interval $j$ and $x_j$ is its geometric midpoint. Ordinary least squares assigns equal weight to the five interval means. The slope $\beta$ measures the change in CR, in percentage points, for each doubling of byte position. We exclude positions below 1~KiB, where little preceding context is available. Table~\ref{tab:long-context} reports these slopes.

Among the representative models at 7--8B, Falcon-H1 improves most rapidly, followed by Qwen3 and RWKV7-G1J, with slopes of $-0.454$, $-0.426$, and $-0.314$, respectively. Their CR decreases by 1.807, 1.700, and 1.266 percentage points between the first and last fitted intervals. This ordering holds in all four categories; each representative model improves most rapidly on mathematics and most slowly on computer science.

Relative to Qwen3-8B-Base, Falcon-H1-7B shifts from 0.025 CR points higher in the 1--2~KiB interval to 0.082 points lower at 16--32~KiB. RWKV7-G1J-7.2B shifts from 0.067 points lower to 0.367 points higher. RWKV's short-context advantage therefore decays as document position grows, even though its absolute CR continues to improve. Our observation is consistent with von Oswald et al.~\cite{vonoswald2026mesanet}, who find that all evaluated linear language models outperform attention-based models at early sequence positions but gradually lose this advantage as the sequence grows.

To examine the role of training context, we additionally evaluate older RWKV7-G1A-1.5B/2.9B and RWKV7-G0A2-7.2B/G0A3-13.3B checkpoints trained with 4,096-token contexts, alongside matched-size RWKV7-G1J checkpoints trained with 16,384 tokens. The short-to-long increase in RWKV's CR gap relative to Qwen3 is smaller for G1J (0.33--0.55 percentage points) than for the older checkpoints (0.41--0.65), indicating that longer training contexts reduce relative decay in these comparisons.

\FloatBarrier
\subsection{Scaling Laws for Compression Rate}

We fit CR as a power law in total parameter count $P$ (in billions) with an additive constant, following a common scaling-law form \cite{henighan2020scaling}:
\begin{equation}
\mathrm{CR}(P)=aP^b+c.
\label{eq:scaling-law}
\end{equation}
We estimate $a$, $b$, and $c$ by minimizing squared errors. The exponent $b$ controls how quickly CR declines with model size, while $c$ is the asymptotic CR as $P$ increases. For mixture-of-experts models, $P$ counts all experts. We fit the 80 models' byte-weighted CR across all 14 categories and separately for each text type.

\begin{figure}[!htbp]
\centering
\includegraphics[width=\linewidth]{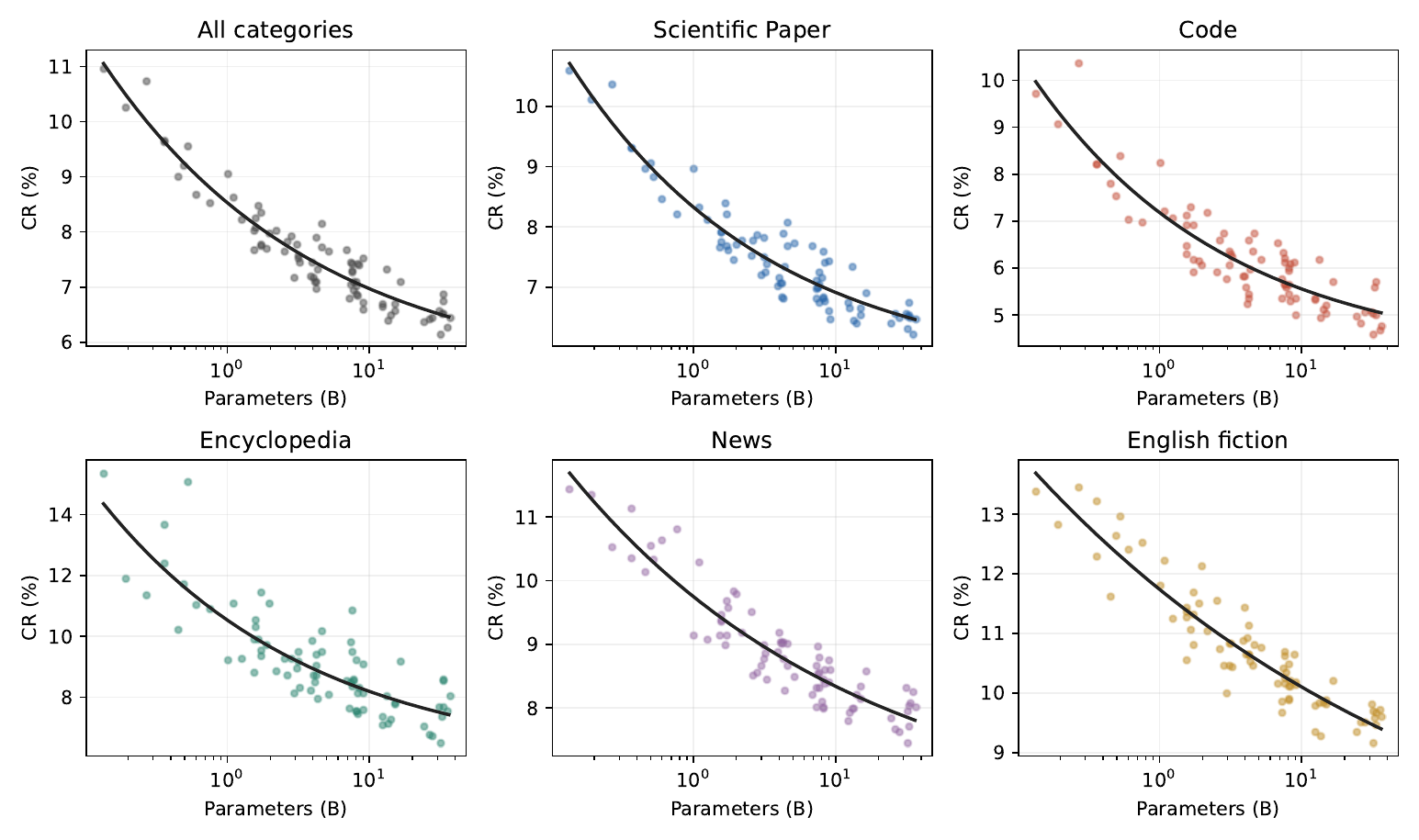}
\caption{CR versus total parameters for 80 models. Each panel shows a power-law-plus-constant fit. For each text type, CR is computed from the total code length and UTF-8 bytes across its categories.}
\label{fig:scaling}
\end{figure}

For the aggregate score, we obtain
\begin{equation}
\widehat{\mathrm{CR}}(P)=3.200P^{-0.290}+5.328,
\qquad R^2=0.915.
\label{eq:aggregate-scaling-fit}
\end{equation}
The fitted curve accounts for 91.5\% of the cross-model CR variation and has a root mean squared error of 0.293 percentage points. The fitted CR curve flattens as model size grows (Figure~\ref{fig:scaling}).

\begin{table}[H]
\centering
\small
\caption{Power-law-plus-constant fits to all 80 models by data type. All types use byte-weighted CR within the included categories.}
\label{tab:group-scaling}
\setlength{\tabcolsep}{5pt}
\begin{tabular}{lrrrrrr}
\toprule
Type & Cats. & $a$ & $b$ & $c$ & $R^2$ & RMSE \\
\midrule
All categories & 14 & 3.200 & -0.290 & 5.328 & 0.915 & 0.293 \\
Scientific Paper & 5 & 2.783 & -0.309 & 5.543 & 0.897 & 0.299 \\
Code & 5 & 3.140 & -0.317 & 4.040 & 0.834 & 0.454 \\
Encyclopedia & 2 & 4.776 & -0.293 & 5.755 & 0.721 & 0.901 \\
News & 1 & 3.624 & -0.214 & 6.124 & 0.871 & 0.323 \\
English fiction & 1 & 5.743 & -0.146 & 5.996 & 0.872 & 0.366 \\
\bottomrule
\end{tabular}
\par\smallskip
\parbox{0.97\linewidth}{\footnotesize Parameters correspond to $\mathrm{CR}(P)=aP^b+c$, with $P$ measured in billions. RMSE is measured across all 80 models in CR percentage points.}
\end{table}

Scaling strength varies by text type: code has the steepest exponent ($-0.317$), while English fiction has the shallowest ($-0.146$). The fits account for 72.1\%--89.7\% of the cross-model variation (Table~\ref{tab:group-scaling}), showing that model size explains substantially more variation for some types than for others.

At the category level, fit quality varies widely: $R^2$ ranges from 0.594 for non-English encyclopedia articles to 0.935 for biology preprints (Appendix~\ref{app:categories}).

We also fit Eq.~\ref{eq:scaling-law} to the parameter--CR Pareto frontier, retaining models for which no evaluated model has at most as many parameters and lower CR. The all-category frontier contains 20 models and follows $\widehat{\mathrm{CR}}(P)=3.244P^{-0.304}+5.016$ ($R^2=0.992$). Across text types, the frontier exponent ranges from $-0.575$ for encyclopedia articles to $-0.244$ for English fiction. These fits describe the best observed CR at each scale within our model pool; the frontier and its fitted asymptote may change as new models are evaluated (Appendix~\ref{app:pareto-scaling}).

\FloatBarrier
\subsection{Correlation with MMLU}

We evaluate all 80 models on zero-shot MMLU. For each question, we choose the answer with the highest next-token logit among A--D and compute accuracy over 14,042 questions. We compare this accuracy with each model's CR pooled across five Scientific Paper and four Code categories.

We measure the association with Spearman correlation. We obtain 95\% confidence intervals by bootstrapping models 10,000 times for the pooled score and 5,000 times for each category.

\begin{figure}[!htbp]
\centering
\includegraphics[width=\linewidth]{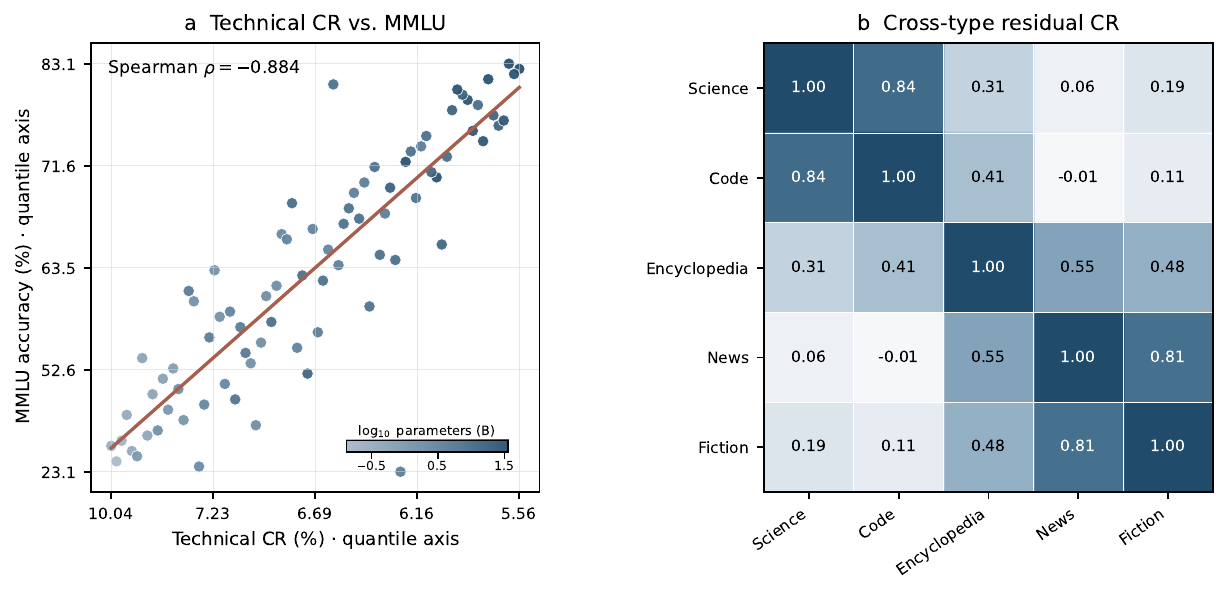}
\caption{Cross-model correlations. (a) Technical-text CR versus zero-shot MMLU accuracy for 80 base models. (b) Pairwise Spearman correlations of residual CR across five text types and the same 80 models, after separate fits against log parameter count.}
\label{fig:mmlu-correlation}
\end{figure}

Across 80 models, lower technical-text CR is associated with higher zero-shot MMLU accuracy (Spearman $\rho=-0.884$, 95\% confidence interval $[-0.942,-0.783]$; Figure~\ref{fig:mmlu-correlation}(a)). The correlation is similar when CR is pooled across all 14 categories ($\rho=-0.871$), and each of the nine technical categories shows the same pattern ($\rho=-0.883$ to $-0.853$).

\FloatBarrier
\subsection{Correlation analyses across text types}

We compute pairwise Pearson correlations among the five text types across 80 models. All ten correlations are positive, ranging from $r=0.803$ to $0.969$. Scientific Paper and Code correlate most strongly ($r=0.969$), followed by News and English fiction ($r=0.963$).

We then fit each type's CR against Overall CR and examine model residuals. A negative residual means better compression on that type than the model's Overall score predicts.The cross-type residual correlations reveal two strongly related pairs: Scientific Paper and Code ($\rho=0.838$), and News and English fiction ($\rho=0.812$; Figure~\ref{fig:mmlu-correlation}(b)). Encyclopedia lies between them, correlating with both pairs but more strongly with News and English fiction ($\rho=0.484$--$0.550$) than with Scientific Paper and Code ($\rho=0.315$--$0.412$). Code and News show almost no residual correlation ($\rho=-0.015$).

\FloatBarrier

\Needspace{10\baselineskip}
\section{Conclusion}
\label{sec:conclusion}

We introduce Uncheatable Eval, a dynamic benchmark that evaluates base language models through lossless compression of newly published text. By converting next-token probabilities into code lengths, it measures predictive ability without task instructions or reference answers. Across 80 models and 14 text categories, CR declines with model size and varies across text types. Our long-context analysis shows that attention-based, hybrid, and recurrent models differ in how their CR changes as more context becomes available. Lower CR on technical text is also strongly associated with higher zero-shot MMLU accuracy. These results support compression of newly published text as a direct way to compare base models across domains.

\section{Limitations}
\label{sec:limitations}

Although newly published text reduces contamination risk, we cannot guarantee its absence from training: publication dates and model training cutoffs may be incomplete, and models may have seen copies of the same content. Our raw-text prediction protocol also limits evaluation to base models, because compression rate does not capture the instruction-following abilities acquired through post-training. Moreover, this contamination risk grows as the dataset ages: new models may train on its public texts, so we must refresh the evaluation data regularly.

\begingroup
\small
\bibliographystyle{unsrt}
\bibliography{references}

@inproceedings{chen2025benchmarking,
  title={Benchmarking Large Language Models Under Data Contamination: A Survey from Static to Dynamic Evaluation},
  author={Chen, Simin and Chen, Yiming and Li, Zexin and Jiang, Yifan and Wan, Zhongwei and He, Yixin and Ran, Dezhi and Gu, Tianle and Li, Haizhou and Xie, Tao and Ray, Baishakhi},
  booktitle={Proceedings of the 2025 Conference on Empirical Methods in Natural Language Processing},
  pages={10080--10098},
  year={2025},
  publisher={Association for Computational Linguistics},
  doi={10.18653/v1/2025.emnlp-main.511},
  url={https://aclanthology.org/2025.emnlp-main.511/}
}

@misc{xu2024benchmark,
  title={Benchmark Data Contamination of Large Language Models: A Survey},
  author={Xu, Cheng and Guan, Shuhao and Greene, Derek and Kechadi, M-Tahar},
  year={2024},
  eprint={2406.04244},
  archivePrefix={arXiv},
  primaryClass={cs.CL},
  url={https://arxiv.org/abs/2406.04244}
}

@inproceedings{white2025livebench,
  title={LiveBench: A Challenging, Contamination-Limited LLM Benchmark},
  author={White, Colin and Dooley, Samuel and Roberts, Manley and Pal, Arka and Feuer, Benjamin and Jain, Siddhartha and Shwartz-Ziv, Ravid and Jain, Neel and Saifullah, Khalid and Dey, Sreemanti and Shubh-Agrawal and Singh Sandha, Sandeep and Naidu, Siddartha and Hegde, Chinmay and LeCun, Yann and Goldstein, Tom and Neiswanger, Willie and Goldblum, Micah},
  booktitle={International Conference on Learning Representations},
  year={2025},
  url={https://proceedings.iclr.cc/paper_files/paper/2025/hash/e4a46394ba5378b3f9a186a5b4c650d1-Abstract-Conference.html}
}

@misc{jain2024livecodebench,
  title={LiveCodeBench: Holistic and Contamination Free Evaluation of Large Language Models for Code},
  author={Jain, Naman and Han, King and Gu, Alex and Li, Wen-Ding and Yan, Fanjia and Zhang, Tianjun and Wang, Sida I. and Solar-Lezama, Armando and Sen, Koushik and Stoica, Ion},
  year={2024},
  eprint={2403.07974},
  archivePrefix={arXiv},
  primaryClass={cs.SE},
  url={https://arxiv.org/abs/2403.07974}
}

@inproceedings{zhu2025freshbench,
  title={Is Your LLM Outdated? A Deep Look at Temporal Generalization},
  author={Zhu, Chenghao and Chen, Nuo and Gao, Yufei and Zhang, Yunyi and Tiwari, Prayag and Wang, Benyou},
  booktitle={Proceedings of the 2025 Conference of the Nations of the Americas Chapter of the Association for Computational Linguistics: Human Language Technologies (Volume 1: Long Papers)},
  pages={7433--7457},
  year={2025},
  publisher={Association for Computational Linguistics},
  doi={10.18653/v1/2025.naacl-long.381},
  url={https://aclanthology.org/2025.naacl-long.381/}
}

@inproceedings{magnusson2024paloma,
  title={Paloma: A Benchmark for Evaluating Language Model Fit},
  author={Magnusson, Ian and Bhagia, Akshita and Hofmann, Valentin and Soldaini, Luca and Jha, Ananya Harsh and Tafjord, Oyvind and Schwenk, Dustin and Walsh, Evan Pete and Elazar, Yanai and Lo, Kyle and Groeneveld, Dirk and Beltagy, Iz and Hajishirzi, Hannaneh and Smith, Noah A. and Richardson, Kyle and Dodge, Jesse},
  booktitle={Advances in Neural Information Processing Systems 37},
  year={2024},
  url={https://proceedings.neurips.cc/paper_files/paper/2024/hash/760b2d94398aa61468aa3bc11506d9ea-Abstract-Datasets_and_Benchmarks_Track.html}
}

@inproceedings{deng2024investigating,
  title={Investigating Data Contamination in Modern Benchmarks for Large Language Models},
  author={Deng, Chunyuan and Zhao, Yilun and Tang, Xiangru and Gerstein, Mark and Cohan, Arman},
  booktitle={Proceedings of the 2024 Conference of the North American Chapter of the Association for Computational Linguistics: Human Language Technologies},
  pages={8706--8719},
  year={2024},
  publisher={Association for Computational Linguistics},
  doi={10.18653/v1/2024.naacl-long.482},
  url={https://aclanthology.org/2024.naacl-long.482/}
}

@inproceedings{dong2024generalization,
  title={Generalization or Memorization: Data Contamination and Trustworthy Evaluation for Large Language Models},
  author={Dong, Yihong and Jiang, Xue and Liu, Huanyu and Jin, Zhi and Gu, Bin and Yang, Mengfei and Li, Ge},
  booktitle={Findings of the Association for Computational Linguistics: ACL 2024},
  pages={12039--12050},
  year={2024},
  publisher={Association for Computational Linguistics},
  doi={10.18653/v1/2024.findings-acl.716},
  url={https://aclanthology.org/2024.findings-acl.716/}
}

@inproceedings{li2024opensourcecontamination,
  title={An Open-Source Data Contamination Report for Large Language Models},
  author={Li, Yucheng and Guo, Yunhao and Guerin, Frank and Lin, Chenghua},
  booktitle={Findings of the Association for Computational Linguistics: EMNLP 2024},
  pages={528--541},
  year={2024},
  publisher={Association for Computational Linguistics},
  doi={10.18653/v1/2024.findings-emnlp.30},
  url={https://aclanthology.org/2024.findings-emnlp.30/}
}

@inproceedings{jacovi2023stop,
  title={Stop Uploading Test Data in Plain Text: Practical Strategies for Mitigating Data Contamination by Evaluation Benchmarks},
  author={Jacovi, Alon and Caciularu, Avi and Goldman, Omer and Goldberg, Yoav},
  booktitle={Proceedings of the 2023 Conference on Empirical Methods in Natural Language Processing},
  pages={5075--5084},
  year={2023},
  publisher={Association for Computational Linguistics},
  doi={10.18653/v1/2023.emnlp-main.308},
  url={https://aclanthology.org/2023.emnlp-main.308/}
}

@inproceedings{zhu2024inference,
  title={Inference-Time Decontamination: Reusing Leaked Benchmarks for Large Language Model Evaluation},
  author={Zhu, Qin and Cheng, Qinyuan and Peng, Runyu and Li, Xiaonan and Peng, Ru and Liu, Tengxiao and Qiu, Xipeng and Huang, Xuanjing},
  booktitle={Findings of the Association for Computational Linguistics: EMNLP 2024},
  pages={9113--9129},
  year={2024},
  publisher={Association for Computational Linguistics},
  doi={10.18653/v1/2024.findings-emnlp.532},
  url={https://aclanthology.org/2024.findings-emnlp.532/}
}

@inproceedings{wu2025antileakbench,
  title={AntiLeakBench: Preventing Data Contamination by Automatically Constructing Benchmarks with Updated Real-World Knowledge},
  author={Wu, Xiaobao and Pan, Liangming and Xie, Yuxi and Zhou, Ruiwen and Zhao, Shuai and Ma, Yubo and Du, Mingzhe and Mao, Rui and Luu, Anh Tuan and Wang, William Yang},
  booktitle={Proceedings of the 63rd Annual Meeting of the Association for Computational Linguistics},
  pages={18403--18419},
  year={2025},
  publisher={Association for Computational Linguistics},
  doi={10.18653/v1/2025.acl-long.901},
  url={https://aclanthology.org/2025.acl-long.901/}
}

@inproceedings{xu2025dcr,
  title={DCR: Quantifying Data Contamination in LLMs Evaluation},
  author={Xu, Cheng and Yan, Nan and Guan, Shuhao and Jin, Changhong and Mei, Yuke and Guo, Yibing and Kechadi, Tahar},
  booktitle={Proceedings of the 2025 Conference on Empirical Methods in Natural Language Processing},
  pages={23002--23020},
  year={2025},
  publisher={Association for Computational Linguistics},
  doi={10.18653/v1/2025.emnlp-main.1173},
  url={https://aclanthology.org/2025.emnlp-main.1173/}
}

@inproceedings{fu2025contamination,
  title={Does Data Contamination Detection Work (Well) for LLMs? A Survey and Evaluation on Detection Assumptions},
  author={Fu, Yujuan Velvin and Uzuner, Ozlem and Yetisgen, Meliha and Xia, Fei},
  booktitle={Findings of the Association for Computational Linguistics: NAACL 2025},
  pages={5250--5271},
  year={2025},
  publisher={Association for Computational Linguistics},
  doi={10.18653/v1/2025.findings-naacl.291},
  url={https://aclanthology.org/2025.findings-naacl.291/}
}

@inproceedings{zhang2026controllable,
  title={Controllable Contamination Detection for Reliable LLM Evaluation with Statistical Guarantees},
  author={Zhang, Zheng and Liu, Qi and Liang, Siyuan and Li, Ning and Hu, Zirui and Gao, Weibo and Li, Rui and Huang, Zhenya and Rutkowski, Leszek and Yu, Baosheng and Tao, Dacheng},
  booktitle={Proceedings of the 64th Annual Meeting of the Association for Computational Linguistics},
  pages={30122--30143},
  year={2026},
  publisher={Association for Computational Linguistics},
  doi={10.18653/v1/2026.acl-long.1390},
  url={https://aclanthology.org/2026.acl-long.1390/}
}

@inproceedings{sainz2023nlp,
  title={NLP Evaluation in Trouble: On the Need to Measure LLM Data Contamination for Each Benchmark},
  author={Sainz, Oscar and Campos, Jon and Garc{\'i}a-Ferrero, Iker and Etxaniz, Julen and de Lacalle, Oier Lopez and Agirre, Eneko},
  booktitle={Findings of the Association for Computational Linguistics: EMNLP 2023},
  pages={10776--10787},
  year={2023},
  publisher={Association for Computational Linguistics},
  doi={10.18653/v1/2023.findings-emnlp.722},
  url={https://aclanthology.org/2023.findings-emnlp.722/}
}

@misc{singh2024contamination,
  title={Evaluation Data Contamination in LLMs: How Do We Measure It and (When) Does It Matter?},
  author={Singh, Aaditya K. and Kocyigit, Muhammed Yusuf and Poulton, Andrew and Esiobu, David and Lomeli, Maria and Szilvasy, Gergely and Hupkes, Dieuwke},
  year={2024},
  eprint={2411.03923},
  archivePrefix={arXiv},
  primaryClass={cs.CL},
  url={https://arxiv.org/abs/2411.03923}
}

@inproceedings{xu2026livefact,
  title={{LiveFact}: A Dynamic, Time-Aware Benchmark for {LLM}-Driven Fake News Detection},
  author={Xu, Cheng and Jin, Changhong and Niu, Yingjie and Yan, Nan and Mei, Yuke and Guan, Shuhao and Chen, Liming and Kechadi, M-Tahar},
  booktitle={Proceedings of the 64th Annual Meeting of the Association for Computational Linguistics (Volume 1: Long Papers)},
  pages={11881--11910},
  year={2026},
  publisher={Association for Computational Linguistics},
  doi={10.18653/v1/2026.acl-long.546},
  url={https://aclanthology.org/2026.acl-long.546/}
}

@inproceedings{yu2024kieval,
  title={{KIE}val: A Knowledge-grounded Interactive Evaluation Framework for Large Language Models},
  author={Yu, Zhuohao and Gao, Chang and Yao, Wenjin and Wang, Yidong and Ye, Wei and Wang, Jindong and Xie, Xing and Zhang, Yue and Zhang, Shikun},
  booktitle={Proceedings of the 62nd Annual Meeting of the Association for Computational Linguistics (Volume 1: Long Papers)},
  pages={5967--5985},
  year={2024},
  publisher={Association for Computational Linguistics},
  doi={10.18653/v1/2024.acl-long.325},
  url={https://aclanthology.org/2024.acl-long.325/}
}

@inproceedings{qian2024varbench,
  title={{VarBench}: Robust Language Model Benchmarking Through Dynamic Variable Perturbation},
  author={Qian, Kun and Wan, Shunji and Tang, Claudia and Wang, Youzhi and Zhang, Xuanming and Chen, Maximillian and Yu, Zhou},
  booktitle={Findings of the Association for Computational Linguistics: EMNLP 2024},
  pages={16131--16161},
  year={2024},
  publisher={Association for Computational Linguistics},
  doi={10.18653/v1/2024.findings-emnlp.946},
  url={https://aclanthology.org/2024.findings-emnlp.946/}
}

@inproceedings{hidayat2025simulating,
  title={Simulating Training Data Leakage in Multiple-Choice Benchmarks for {LLM} Evaluation},
  author={Hidayat, Naila Shafirni and Al Kautsar, Muhammad Dehan and Wicaksono, Alfan Farizki and Koto, Fajri},
  booktitle={Proceedings of the 5th Workshop on Evaluation and Comparison of NLP Systems},
  pages={21--39},
  year={2025},
  publisher={Association for Computational Linguistics},
  doi={10.18653/v1/2025.eval4nlp-1.3},
  url={https://aclanthology.org/2025.eval4nlp-1.3/}
}

@inproceedings{chai2026benchmarks,
  title={When Benchmarks Leak: Inference-Time Decontamination for {LLM}s},
  author={Chai, Jianzhe and Yu, Zhe and Sakuma, Jun},
  booktitle={Proceedings of the 64th Annual Meeting of the Association for Computational Linguistics (Volume 1: Long Papers)},
  pages={44743--44760},
  year={2026},
  publisher={Association for Computational Linguistics},
  doi={10.18653/v1/2026.acl-long.2071},
  url={https://aclanthology.org/2026.acl-long.2071/}
}

@misc{xiao2026leaderboards,
  title={Contamination Inflates Scores but Rarely Reorders Large Language Model Leaderboards},
  author={Xiao, Xingyao and Cheng, Yihong},
  year={2026},
  eprint={2609.02899},
  archivePrefix={arXiv},
  primaryClass={cs.CL},
  url={https://arxiv.org/abs/2609.02899}
}

@inproceedings{kim2026oaks,
  title={Can Large Language Models Keep Up? Benchmarking Online Adaptation to Continual Knowledge Streams},
  author={Kim, Jiyeon and Lee, Hyunji and Zhou, Dylan and Park, Sue Hyun and Yoon, Seunghyun and Bui, Trung and Dernoncourt, Franck and Cha, Sungmin and Seo, Minjoon},
  booktitle={Proceedings of the 64th Annual Meeting of the Association for Computational Linguistics (Volume 1: Long Papers)},
  pages={42240--42272},
  year={2026},
  publisher={Association for Computational Linguistics},
  doi={10.18653/v1/2026.acl-long.1956},
  url={https://aclanthology.org/2026.acl-long.1956/}
}

@inproceedings{liu2026rag,
  title={{RAG} or Learning? Understanding the Limits of {LLM} Adaptation under Continuous Knowledge Drift in the Real World},
  author={Liu, Hanbing and Cao, Lang and Li, Yang},
  booktitle={Findings of the Association for Computational Linguistics: ACL 2026},
  pages={11234--11252},
  year={2026},
  publisher={Association for Computational Linguistics},
  doi={10.18653/v1/2026.findings-acl.546},
  url={https://aclanthology.org/2026.findings-acl.546/}
}

@misc{allawati2026contamination,
  title={LLM Benchmark Datasets Should Be Contamination-Resistant},
  author={Al-Lawati, Ali and Lucas, Jason and Lee, Dongwon and Wang, Suhang},
  year={2026},
  eprint={2605.19999},
  archivePrefix={arXiv},
  primaryClass={cs.LG},
  note={Accepted to ICML 2026 Position Paper Track},
  url={https://arxiv.org/abs/2605.19999}
}

@misc{li2024compression,
  title={Evaluating Large Language Models for Generalization and Robustness via Data Compression},
  author={Li, Yucheng and Guo, Yunhao and Guerin, Frank and Lin, Chenghua},
  year={2024},
  eprint={2402.00861},
  archivePrefix={arXiv},
  primaryClass={cs.CL},
  url={https://arxiv.org/abs/2402.00861}
}

@inproceedings{huang2024compression,
  title={Compression Represents Intelligence Linearly},
  author={Huang, Yuzhen and Zhang, Jinghan and Shan, Zifei and He, Junxian},
  booktitle={First Conference on Language Modeling},
  year={2024},
  url={https://openreview.net/forum?id=SHMj84U5SH}
}

@misc{deletang2023compression,
  title={Language Modeling Is Compression},
  author={Del{\'e}tang, Gr{\'e}goire and Ruoss, Anian and Duquenne, Paul-Ambroise and Catt, Elliot and Genewein, Tim and Mattern, Christopher and Grau-Moya, Jordi and Wenliang, Li Kevin and Aitchison, Matthew and Orseau, Laurent and Hutter, Marcus and Veness, Joel},
  year={2023},
  eprint={2309.10668},
  archivePrefix={arXiv},
  primaryClass={cs.LG},
  url={https://arxiv.org/abs/2309.10668}
}

@misc{roberts2023time,
  title={Data Contamination Through the Lens of Time},
  author={Roberts, Manley and Thakur, Himanshu and Herlihy, Christine and White, Colin and Dooley, Samuel},
  year={2023},
  eprint={2310.10628},
  archivePrefix={arXiv},
  primaryClass={cs.CL},
  url={https://arxiv.org/abs/2310.10628}
}

@misc{peng2025rwkv7,
  title={{RWKV}-7 ``Goose'' with Expressive Dynamic State Evolution},
  author={Peng, Bo and Zhang, Ruichong and Goldstein, Daniel and others},
  year={2025},
  eprint={2503.14456},
  archivePrefix={arXiv},
  primaryClass={cs.CL},
  url={https://arxiv.org/abs/2503.14456}
}

@misc{zuo2025falconh1,
  title={{Falcon-H1}: A Family of Hybrid-Head Language Models Redefining Efficiency and Performance},
  author={Zuo, Jingwei and Velikanov, Maksim and Chahed, Ilyas and others},
  year={2025},
  eprint={2507.22448},
  archivePrefix={arXiv},
  primaryClass={cs.CL},
  url={https://arxiv.org/abs/2507.22448}
}

@misc{yang2025qwen3,
  title={{Qwen3} Technical Report},
  author={Yang, An and Li, Anfeng and Yang, Baosong and others},
  year={2025},
  eprint={2505.09388},
  archivePrefix={arXiv},
  primaryClass={cs.CL},
  url={https://arxiv.org/abs/2505.09388}
}

@misc{qwen2026qwen35,
  title={{Qwen3.5}: Towards Native Multimodal Agents},
  author={{Qwen Team}},
  year={2026},
  month={feb},
  howpublished={Official model release page},
  url={https://qwen.ai/blog?id=qwen3.5}
}

@misc{gemmateam2026gemma4,
  title={Gemma 4 Technical Report},
  author={{Gemma Team} and El Abd, Sherif and Aggarwal, Vaibhav and others},
  year={2026},
  eprint={2607.02770},
  archivePrefix={arXiv},
  primaryClass={cs.LG},
  url={https://arxiv.org/abs/2607.02770}
}

@misc{nvidia2025nemotron3nano,
  title={Nemotron 3 Nano: Open, Efficient Mixture-of-Experts Hybrid Mamba-Transformer Model for Agentic Reasoning},
  author={{NVIDIA} and Blakeman, Aaron and Grattafiori, Aaron and others},
  year={2025},
  eprint={2512.20848},
  archivePrefix={arXiv},
  primaryClass={cs.LG},
  url={https://arxiv.org/abs/2512.20848}
}

@misc{henighan2020scaling,
  title={Scaling Laws for Autoregressive Generative Modeling},
  author={Henighan, Tom and Kaplan, Jared and Katz, Mor and Chen, Mark and Hesse, Christopher and Jackson, Jacob and others},
  year={2020},
  eprint={2010.14701},
  archivePrefix={arXiv},
  primaryClass={cs.LG},
  url={https://arxiv.org/abs/2010.14701}
}

@inproceedings{vonoswald2026mesanet,
  title={MesaNet: Sequence Modeling by Locally Optimal Test-Time Training},
  author={von Oswald, Johannes and Scherrer, Nino and Kobayashi, Seijin and Versari, Luca and Yang, Songlin and Mittal, Sarthak and Schlegel, Maximilian and Maile, Kaitlin and Schimpf, Yanick and Sieberling, Oliver and Meulemans, Alexander and Saurous, Rif A. and Lajoie, Guillaume and Frenkel, Charlotte and Pascanu, Razvan and Ag{\"u}era y Arcas, Blaise and Sacramento, Jo{\~a}o},
  booktitle={International Conference on Learning Representations},
  year={2026},
  url={https://proceedings.iclr.cc/paper_files/paper/2026/file/2c3d5a4eff20d24390a445762d61bfbb-Abstract-Conference.html}
}
\endgroup

\clearpage
\appendix
\begingroup
\section{Complete Model-by-Category Results}
\label{app:complete-results}
Table~\ref{tab:complete-results} reports category-level results for the 80-model cohort used in the main cross-model analyses.
\par\medskip
\fontsize{7}{8.6}\selectfont
\setlength{\tabcolsep}{1.5pt}
\renewcommand{\arraystretch}{1.04}
\begin{longtable}{@{}>{\raggedright\arraybackslash}p{3.4cm}>{\raggedleft\arraybackslash}p{0.95cm}*{14}{>{\raggedleft\arraybackslash}p{0.64cm}}>{\raggedleft\arraybackslash}p{0.95cm}@{}}
\caption{July 2026 category-level results for the 80-model analysis cohort. All score cells are compression rate (CR, \%); lower is better.}\label{tab:complete-results}\\
\toprule
\multirow{2}{*}{\raggedright Model} & \multirow{2}{*}{\raggedleft Params.} & \multicolumn{4}{c}{General text} & \multicolumn{5}{c}{Scientific Paper} & \multicolumn{5}{c}{Code} & Overall \\
\cmidrule(lr){3-6}\cmidrule(lr){7-11}\cmidrule(lr){12-16}\cmidrule(lr){17-17}
& & EF & News & EE & NEE & Bio & CS & Math & Sci-O & Phys & C++ & JS & MD & Other & Py & Overall \\
\midrule
\endfirsthead
\caption[]{July 2026 category-level results for the 80-model analysis cohort. All score cells are compression rate (CR, \%); lower is better. (continued)}\\
\toprule
\multirow{2}{*}{\raggedright Model} & \multirow{2}{*}{\raggedleft Params.} & \multicolumn{4}{c}{General text} & \multicolumn{5}{c}{Scientific Paper} & \multicolumn{5}{c}{Code} & Overall \\
\cmidrule(lr){3-6}\cmidrule(lr){7-11}\cmidrule(lr){12-16}\cmidrule(lr){17-17}
& & EF & News & EE & NEE & Bio & CS & Math & Sci-O & Phys & C++ & JS & MD & Other & Py & Overall \\
\midrule
\endhead
\midrule
\multicolumn{17}{r}{\emph{Continued on next page}} \\
\endfoot
\bottomrule
\endlastfoot
\midrule
\multicolumn{17}{@{}l}{\textbf{>20B}} \\
\addlinespace[1.5pt]
Gemma-4-31B & 31.3 & \textbf{9.17} & \textbf{7.46} & \textbf{6.98} & \textbf{6.09} & \textbf{6.43} & 6.59 & 5.98 & 6.16 & 6.31 & \textbf{3.16} & \textbf{3.70} & \textbf{7.86} & \textbf{3.98} & \textbf{4.15} & \textbf{6.00} \\
Qwen3.5-35B-A3B-Base & 34.7 & 9.73 & 8.25 & 7.61 & 7.49 & \textbf{6.43} & \textbf{6.52} & \textbf{5.78} & \textbf{6.09} & \textbf{6.24} & 3.33 & 3.75 & 7.92 & 4.04 & 4.24 & 6.24 \\
Gemma-4-26B-A4B & 25.8 & 9.52 & 7.68 & 7.22 & 6.41 & 6.62 & 6.92 & 6.22 & 6.44 & 6.56 & 3.29 & 3.87 & 8.34 & 4.17 & 4.38 & 6.26 \\
Mistral-Small-24B & 24.0 & 9.36 & 7.85 & 7.34 & 6.82 & 6.44 & 6.79 & 6.08 & 6.30 & 6.33 & 3.44 & 4.04 & 8.39 & 4.30 & 4.49 & 6.28 \\
Gemma-3-27B & 27.4 & 9.53 & 7.64 & 7.15 & 6.35 & 6.62 & 6.90 & 6.07 & 6.39 & 6.44 & 3.46 & 4.08 & 8.58 & 4.41 & 4.61 & 6.30 \\
Seed-OSS-36B-Base & 36.2 & 9.61 & 8.02 & 7.48 & 8.51 & 6.52 & 6.83 & 5.97 & 6.36 & 6.57 & 3.26 & 3.89 & 8.12 & 4.07 & 4.28 & 6.39 \\
Nemotron-3-Nano-30B-A3B & 31.6 & 9.70 & 7.96 & 7.33 & 7.40 & 6.60 & 6.84 & 6.19 & 6.43 & 6.49 & 3.44 & 4.15 & 8.49 & 4.44 & 4.50 & 6.43 \\
Qwen2.5-32B & 32.8 & 9.69 & 8.10 & 7.74 & 7.64 & 6.62 & 6.96 & 5.84 & 6.40 & 6.52 & 3.41 & 4.11 & 8.55 & 4.29 & 4.47 & 6.45 \\
Qwen3-30B-A3B-Base & 30.5 & 9.82 & 8.33 & 7.86 & 7.57 & 6.69 & 7.04 & 5.90 & 6.47 & 6.54 & 3.41 & 4.07 & 8.76 & 4.35 & 4.53 & 6.52 \\
OLMo-3-32B & 32.2 & 9.60 & 8.05 & 7.71 & 9.24 & 6.58 & 6.87 & 6.18 & 6.42 & 6.60 & 4.09 & 4.85 & 8.88 & 5.07 & 4.91 & 6.79 \\
marin-32b-base & 32.5 & 9.47 & 7.72 & 7.40 & 9.63 & 6.78 & 7.23 & 6.33 & 6.63 & 6.74 & 4.13 & 4.90 & 9.18 & 5.02 & 5.11 & 6.88 \\
\midrule
\multicolumn{17}{@{}l}{\textbf{$\sim$14B}} \\
\addlinespace[1.5pt]
RWKV7-G1J-13.3B & 13.3 & \textbf{9.30} & 8.01 & \textbf{7.45} & 6.87 & 6.62 & \textbf{6.68} & 6.13 & \textbf{6.32} & 6.48 & 3.59 & \textbf{4.07} & \textbf{8.14} & \textbf{4.30} & \textbf{4.42} & \textbf{6.31} \\
Ministral-3-14B & 13.9 & 9.86 & 8.22 & 7.64 & 6.99 & \textbf{6.44} & 6.80 & 6.03 & 6.33 & \textbf{6.36} & 3.55 & 4.21 & 8.60 & 4.46 & 4.65 & 6.44 \\
Qwen3-14B-Base & 14.8 & 9.89 & 8.35 & 7.90 & 7.66 & 6.68 & 6.98 & \textbf{5.93} & 6.45 & 6.55 & \textbf{3.44} & 4.10 & 8.62 & 4.35 & 4.52 & 6.53 \\
Mistral-Nemo-Base-2407 & 12.2 & 9.36 & 7.93 & 7.51 & 7.23 & 6.59 & 7.12 & 6.33 & 6.57 & 6.65 & 3.75 & 4.41 & 8.98 & 4.66 & 4.86 & 6.57 \\
gemma-3-12b-pt & 12.2 & 9.81 & \textbf{7.81} & \textbf{7.45} & \textbf{6.83} & 6.85 & 7.13 & 6.35 & 6.63 & 6.67 & 3.68 & 4.32 & 8.94 & 4.64 & 4.82 & 6.57 \\
Qwen2.5-14B & 14.8 & 9.82 & 8.15 & 7.84 & 7.83 & 6.77 & 7.14 & 6.03 & 6.57 & 6.69 & 3.59 & 4.31 & 8.85 & 4.48 & 4.66 & 6.62 \\
Gravity-16B-A3B-Base & 16.2 & 10.22 & 8.58 & 8.38 & 9.88 & 6.82 & 7.25 & 6.62 & 6.78 & 7.01 & 4.02 & 4.88 & 9.25 & 5.05 & 5.17 & 7.14 \\
Llama-2-13b-hf & 13.0 & 9.85 & 7.99 & 7.71 & 8.38 & 7.26 & 7.79 & 7.07 & 7.22 & 7.31 & 4.37 & 5.24 & 10.02 & 5.37 & 5.62 & 7.23 \\
\midrule
\multicolumn{17}{@{}l}{\textbf{$\sim$7B}} \\
\addlinespace[1.5pt]
Qwen3.5-9B-Base & 9.0 & 10.15 & 8.60 & 8.02 & 8.24 & 6.67 & \textbf{6.76} & \textbf{6.06} & \textbf{6.34} & \textbf{6.50} & 3.65 & \textbf{4.08} & \textbf{8.28} & \textbf{4.33} & \textbf{4.51} & \textbf{6.59} \\
Ministral-3-8B & 8.9 & 10.20 & 8.42 & 7.90 & 7.34 & \textbf{6.64} & 7.02 & 6.25 & 6.54 & 6.58 & 3.71 & 4.42 & 8.93 & 4.66 & 4.84 & 6.68 \\
RWKV7-G1J-7.2B & 7.2 & \textbf{9.69} & 8.34 & 7.84 & 7.48 & 6.94 & 7.05 & 6.54 & 6.67 & 6.84 & 3.91 & 4.44 & 8.75 & 4.67 & 4.81 & 6.71 \\
Meta-Llama-3.1-8B & 8.0 & 9.89 & \textbf{8.01} & \textbf{7.66} & 7.45 & 6.83 & 7.35 & 6.36 & 6.73 & 6.77 & 3.94 & 4.73 & 9.39 & 4.88 & 5.08 & 6.79 \\
Qwen3-8B-Base & 8.2 & 10.21 & 8.61 & 8.19 & 8.12 & 6.92 & 7.22 & 6.15 & 6.68 & 6.80 & \textbf{3.63} & 4.33 & 8.99 & 4.57 & 4.74 & 6.80 \\
gemma-4-E4B & 7.9 & 10.13 & 8.11 & 7.90 & 7.24 & 7.24 & 7.43 & 6.90 & 7.00 & 7.20 & 3.82 & 4.44 & 9.10 & 4.71 & 4.92 & 6.87 \\
Falcon-H1-7B-Base & 7.6 & 10.16 & 8.33 & 8.00 & 9.09 & 6.88 & 7.23 & 6.17 & 6.65 & 6.71 & 3.96 & 4.69 & 9.41 & 4.90 & 5.12 & 6.95 \\
Mistral-7B-v0.1 & 7.2 & 9.87 & 8.03 & 7.78 & 9.14 & 6.94 & 7.47 & 6.59 & 6.89 & 6.97 & 4.10 & 4.85 & 9.47 & 4.99 & 5.24 & 7.02 \\
Qwen2.5-7B & 7.6 & 10.34 & 8.56 & 8.33 & 8.70 & 7.17 & 7.53 & 6.37 & 6.94 & 7.10 & 3.86 & 4.65 & 9.42 & 4.81 & 4.96 & 7.05 \\
marin-8b-base & 8.0 & 9.91 & 8.02 & 7.86 & 10.45 & 6.91 & 7.25 & 6.44 & 6.75 & 6.75 & 4.33 & 5.17 & 9.63 & 5.25 & 5.46 & 7.16 \\
Hunyuan-7B-Pretrain & 7.5 & 10.70 & 8.80 & 8.35 & 8.40 & 7.35 & 7.60 & 6.81 & 7.13 & 7.34 & 4.04 & 4.72 & 9.36 & 4.89 & 5.04 & 7.18 \\
Apertus-8B-2509 & 8.1 & 10.48 & 8.46 & 7.95 & \textbf{7.05} & 7.44 & 7.89 & 7.55 & 7.43 & 7.75 & 4.17 & 5.00 & 9.70 & 5.15 & 5.42 & 7.25 \\
Minitron-8B & 8.3 & 10.11 & 8.39 & 8.11 & 8.55 & 7.22 & 7.75 & 7.38 & 7.28 & 7.49 & 4.17 & 5.17 & 10.03 & 5.35 & 5.47 & 7.32 \\
Olmo-Hybrid-7B & 7.4 & 10.27 & 8.55 & 8.27 & 10.60 & 7.03 & 7.34 & 6.64 & 6.87 & 7.09 & 4.61 & 5.46 & 9.74 & 5.65 & 5.40 & 7.39 \\
ZAYA1-base & 8.8 & 10.65 & 8.76 & 8.44 & 9.64 & 7.42 & 7.78 & 7.12 & 7.32 & 7.45 & 4.32 & 5.19 & 10.01 & 5.35 & 5.54 & 7.50 \\
OLMo-3-7B & 7.3 & 10.41 & 8.67 & 8.37 & 11.08 & 7.13 & 7.45 & 6.77 & 6.99 & 7.22 & 4.71 & 5.58 & 9.93 & 5.77 & 5.50 & 7.54 \\
Llama-2-7b-hf & 6.7 & 10.18 & 8.23 & 8.03 & 8.98 & 7.59 & 8.14 & 7.47 & 7.57 & 7.67 & 4.66 & 5.59 & 10.55 & 5.70 & 5.96 & 7.59 \\
Falcon3-7B-Base & 7.5 & 10.64 & 8.98 & 8.67 & 12.74 & 7.07 & 7.48 & 6.52 & 6.89 & 6.98 & 4.24 & 5.13 & 10.29 & 5.24 & 5.53 & 7.60 \\
\pagebreak
\midrule
\multicolumn{17}{@{}l}{\textbf{$\sim$3B}} \\
\addlinespace[1.5pt]
Qwen3.5-4B-Base & 4.2 & 10.65 & 9.05 & 8.49 & 8.95 & 6.99 & \textbf{7.07} & 6.41 & \textbf{6.66} & \textbf{6.83} & 3.98 & 4.43 & 8.79 & 4.66 & 4.83 & \textbf{6.98} \\
Spark-X2.5-4B-Base & 4.1 & 10.94 & 9.03 & 8.49 & 8.54 & 7.24 & 7.31 & 6.62 & 6.94 & 7.16 & 3.96 & \textbf{4.39} & \textbf{8.53} & \textbf{4.56} & \textbf{4.61} & 7.02 \\
RWKV7-G1J-2.9B & 2.9 & \textbf{10.02} & 8.67 & 8.22 & 8.05 & 7.28 & 7.44 & 6.96 & 7.05 & 7.23 & 4.25 & 4.86 & 9.36 & 5.05 & 5.21 & 7.12 \\
Nanbeige4.2-3B-Base & 4.2 & 11.13 & 9.01 & 8.29 & 10.41 & 7.01 & 7.13 & 6.47 & 6.71 & 6.86 & 4.00 & 4.64 & 8.79 & 4.79 & 4.81 & 7.15 \\
Qwen3-4B-Base & 4.0 & 10.66 & 9.03 & 8.63 & 8.79 & 7.24 & 7.52 & \textbf{6.40} & 6.96 & 7.11 & \textbf{3.87} & 4.60 & 9.45 & 4.84 & 4.98 & 7.15 \\
Ministral-3-3B & 3.8 & 10.89 & 8.90 & 8.41 & 8.10 & \textbf{6.98} & 7.44 & 6.69 & 6.94 & 7.00 & 4.08 & 4.87 & 9.62 & 5.08 & 5.25 & 7.16 \\
gemma-3-4b-pt & 4.3 & 10.54 & \textbf{8.28} & \textbf{8.14} & \textbf{7.79} & 7.45 & 7.72 & 6.98 & 7.20 & 7.28 & 4.22 & 4.94 & 9.88 & 5.24 & 5.40 & 7.22 \\
Llama-3.2-3B & 3.2 & 10.45 & 8.46 & 8.25 & 8.37 & 7.38 & 7.89 & 6.97 & 7.28 & 7.34 & 4.43 & 5.31 & 10.23 & 5.43 & 5.59 & 7.38 \\
gemma-4-E2B & 5.1 & 10.77 & 8.51 & 8.38 & 7.84 & 7.72 & 8.03 & 7.55 & 7.55 & 7.79 & 4.44 & 5.17 & 10.09 & 5.41 & 5.60 & 7.49 \\
Nanbeige4-3B-Base & 3.9 & 11.43 & 9.18 & 8.78 & 10.78 & 7.27 & 7.50 & 6.77 & 7.04 & 7.18 & 4.21 & 4.90 & 9.46 & 5.21 & 5.23 & 7.50 \\
Qwen2.5-3B & 3.1 & 10.85 & 9.00 & 8.80 & 9.50 & 7.61 & 7.99 & 6.82 & 7.38 & 7.57 & 4.25 & 5.11 & 10.14 & 5.25 & 5.37 & 7.55 \\
Falcon-H1-3B-Base & 3.1 & 10.84 & 8.87 & 8.65 & 10.25 & 7.44 & 7.72 & 6.64 & 7.12 & 7.22 & 4.52 & 5.27 & 10.27 & 5.53 & 5.68 & 7.57 \\
MiniCPM5-2B-Base & 2.5 & 11.55 & 9.51 & 8.92 & 9.62 & 7.67 & 7.75 & 7.19 & 7.35 & 7.66 & 4.20 & 4.99 & 9.66 & 5.25 & 5.31 & 7.62 \\
SmolLM3-3B-Base & 3.1 & 10.48 & 8.78 & 8.51 & 9.39 & 7.71 & 8.17 & 7.57 & 7.65 & 7.95 & 4.39 & 5.40 & 10.41 & 5.66 & 5.64 & 7.69 \\
Llama-3.1-Minitron-4B-Width & 4.5 & 10.47 & 8.67 & 8.39 & 10.45 & 7.49 & 8.03 & 7.62 & 7.55 & 7.76 & 4.38 & 5.43 & 10.45 & 5.61 & 5.69 & 7.71 \\
gemma-2-2b & 2.6 & 10.74 & 8.53 & 8.59 & 8.82 & 7.62 & 8.28 & 7.54 & 7.67 & 7.77 & 4.60 & 5.45 & 10.84 & 5.79 & 5.98 & 7.73 \\
Minitron-4B-Base & 4.2 & 10.66 & 8.77 & 8.59 & 9.45 & 7.62 & 8.20 & 7.97 & 7.74 & 8.01 & 4.53 & 5.65 & 10.77 & 5.84 & 5.91 & 7.84 \\
stablelm-3b-4e1t & 2.8 & 10.46 & 8.56 & 8.38 & 10.03 & 7.68 & 8.35 & 7.71 & 7.76 & 7.89 & 4.60 & 5.67 & 11.07 & 6.02 & 6.01 & 7.87 \\
Index-1.9B & 2.2 & 11.04 & 9.19 & 8.96 & 8.79 & 7.87 & 8.32 & 7.11 & 7.68 & 7.71 & 4.99 & 6.33 & 11.42 & 6.36 & 6.45 & 8.02 \\
Llama-3.1-Minitron-4B-Depth & 4.5 & 10.82 & 9.02 & 8.77 & 11.41 & 7.87 & 8.42 & 8.05 & 7.93 & 8.18 & 4.65 & 5.77 & 10.98 & 5.96 & 6.03 & 8.13 \\
\midrule
\multicolumn{17}{@{}l}{\textbf{$<2$B}} \\
\addlinespace[1.5pt]
RWKV7-G1J-1.5B & 1.5 & \textbf{10.56} & 9.16 & \textbf{8.76} & \textbf{8.84} & 7.71 & 7.91 & 7.46 & 7.50 & 7.69 & 4.69 & 5.35 & 10.10 & 5.53 & 5.67 & \textbf{7.64} \\
Spark-X2.5-1.7B-Base & 1.7 & 11.69 & 9.68 & 9.19 & 9.56 & 7.82 & 7.95 & 7.21 & 7.54 & 7.76 & 4.48 & \textbf{5.03} & \textbf{9.51} & \textbf{5.18} & \textbf{5.22} & 7.70 \\
Qwen3.5-2B-Base & 1.9 & 11.51 & 9.84 & 9.28 & 10.09 & \textbf{7.62} & \textbf{7.75} & 7.08 & \textbf{7.30} & \textbf{7.51} & 4.67 & 5.14 & 9.82 & 5.35 & 5.52 & 7.75 \\
Qwen3-1.7B-Base & 1.7 & 11.33 & 9.58 & 9.24 & 9.80 & 7.78 & 8.08 & \textbf{6.95} & 7.49 & 7.67 & \textbf{4.32} & 5.15 & 10.29 & 5.37 & 5.49 & 7.75 \\
Qwen2.5-1.5B & 1.5 & 11.36 & 9.39 & 9.27 & 10.46 & 8.03 & 8.41 & 7.22 & 7.77 & 7.99 & 4.56 & 5.49 & 10.74 & 5.61 & 5.72 & 8.00 \\
Youtu-LLM-2B-Base & 2.0 & 12.12 & 9.81 & 9.12 & 12.77 & 7.85 & 7.93 & 7.41 & 7.54 & 7.77 & 4.43 & 5.21 & 9.85 & 5.32 & 5.31 & 8.03 \\
Falcon-H1-1.5B-Deep-Base & 1.6 & 11.27 & 9.36 & 9.18 & 11.33 & 7.90 & 8.24 & 7.10 & 7.58 & 7.83 & 5.07 & 5.88 & 11.06 & 6.16 & 6.20 & 8.16 \\
Llama-3.2-1B & 1.2 & 11.26 & 9.08 & 8.98 & 9.50 & 8.08 & 8.65 & 7.73 & 8.00 & 8.09 & 5.11 & 6.08 & 11.40 & 6.19 & 6.32 & 8.18 \\
Falcon-H1-1.5B-Base & 1.6 & 11.44 & 9.48 & 9.33 & 11.56 & 8.05 & 8.39 & 7.28 & 7.73 & 7.99 & 5.24 & 6.07 & 11.31 & 6.35 & 6.40 & 8.33 \\
SmolLM2-1.7B & 1.7 & 10.82 & 9.15 & 9.15 & 13.40 & 8.12 & 8.65 & 7.86 & 8.04 & 8.36 & 4.85 & 5.82 & 11.36 & 6.08 & 6.13 & 8.41 \\
stablelm-2-1\_6b & 1.6 & 11.06 & \textbf{9.01} & 8.89 & 10.77 & 8.24 & 8.89 & 8.11 & 8.27 & 8.42 & 5.36 & 6.28 & 11.63 & 6.35 & 6.55 & 8.42 \\
Qwen3.5-0.8B-Base & 0.8 & 12.52 & 10.81 & 10.23 & 11.47 & 8.38 & 8.50 & 7.81 & 8.01 & 8.23 & 5.42 & 5.94 & 10.95 & 6.12 & 6.27 & 8.62 \\
MiniCPM5-1B-Base & 1.1 & 12.23 & 10.30 & 9.82 & 12.16 & 8.41 & 8.64 & 7.93 & 8.15 & 8.40 & 5.11 & 6.14 & 11.38 & 6.38 & 6.76 & 8.70 \\
Qwen3-0.6B-Base & 0.6 & 12.41 & 10.65 & 10.30 & 11.66 & 8.65 & 8.91 & 7.74 & 8.29 & 8.53 & 5.01 & 5.96 & 11.53 & 6.15 & 6.28 & 8.72 \\
gemma-3-1b-pt & 1.0 & 11.80 & 9.15 & 9.30 & 9.17 & 8.81 & 9.29 & 9.07 & 8.76 & 8.99 & 6.46 & 7.23 & 12.39 & 7.58 & 7.30 & 8.95 \\
RWKV7-G1D-0.4B & 0.5 & 11.63 & 10.15 & 9.90 & 10.49 & 8.98 & 9.26 & 8.73 & 8.79 & 9.00 & 5.84 & 6.81 & 12.06 & 6.96 & 6.97 & 8.97 \\
Qwen2.5-0.5B & 0.5 & 12.64 & 10.55 & 10.53 & 12.72 & 9.24 & 9.54 & 8.30 & 8.87 & 9.16 & 5.42 & 6.50 & 12.27 & 6.57 & 6.64 & 9.21 \\
ERNIE-4.5-0.3B-Base-PT & 0.4 & 13.22 & 11.15 & 10.95 & 13.63 & 9.29 & 9.74 & 8.92 & 9.17 & 9.35 & 6.18 & 7.16 & 12.72 & 7.26 & 7.47 & 9.73 \\
SmolLM2-360M & 0.4 & 12.30 & 10.36 & 10.40 & 16.50 & 9.23 & 9.76 & 8.97 & 9.11 & 9.47 & 5.98 & 7.01 & 13.13 & 7.37 & 7.26 & 9.77 \\
Falcon-H1-0.5B-Base & 0.5 & 12.96 & 10.35 & 10.53 & 18.95 & 9.00 & 9.29 & 8.18 & 8.58 & 8.92 & 6.20 & 7.27 & 13.10 & 7.54 & 7.47 & 9.88 \\
RWKV7-G1D-0.1B & 0.2 & 12.82 & 11.35 & 11.19 & 12.50 & 10.16 & 10.40 & 9.93 & 9.90 & 10.19 & 6.98 & 8.05 & 13.65 & 8.23 & 8.14 & 10.25 \\
gemma-3-270m & 0.3 & 13.45 & 10.54 & 10.96 & 11.70 & 10.34 & 10.78 & 10.21 & 10.15 & 10.33 & 8.37 & 9.45 & 14.79 & 10.00 & 8.83 & 10.71 \\
SmolLM2-135M & 0.1 & 13.38 & 11.44 & 11.44 & 18.72 & 10.37 & 10.97 & 10.53 & 10.36 & 10.72 & 6.93 & 8.39 & 14.58 & 9.44 & 8.77 & 11.15 \\
\end{longtable}
\vspace{-0.5\baselineskip}
\noindent\parbox{\linewidth}{\footnotesize Abbreviations follow Table~\ref{tab:main-results}. Overall is the macro-average of the 14 category CRs. Rows are grouped by total parameter count and sorted by unrounded Overall within each group. Bold marks the best displayed value in each group, including ties at the shown precision. Params. is total parameters in billions.}
\endgroup

\clearpage
\section{Long-Context Results across Model Sizes}
\label{app:long-context-scale}
\begin{figure}[H]
\centering
\includegraphics[width=\linewidth]{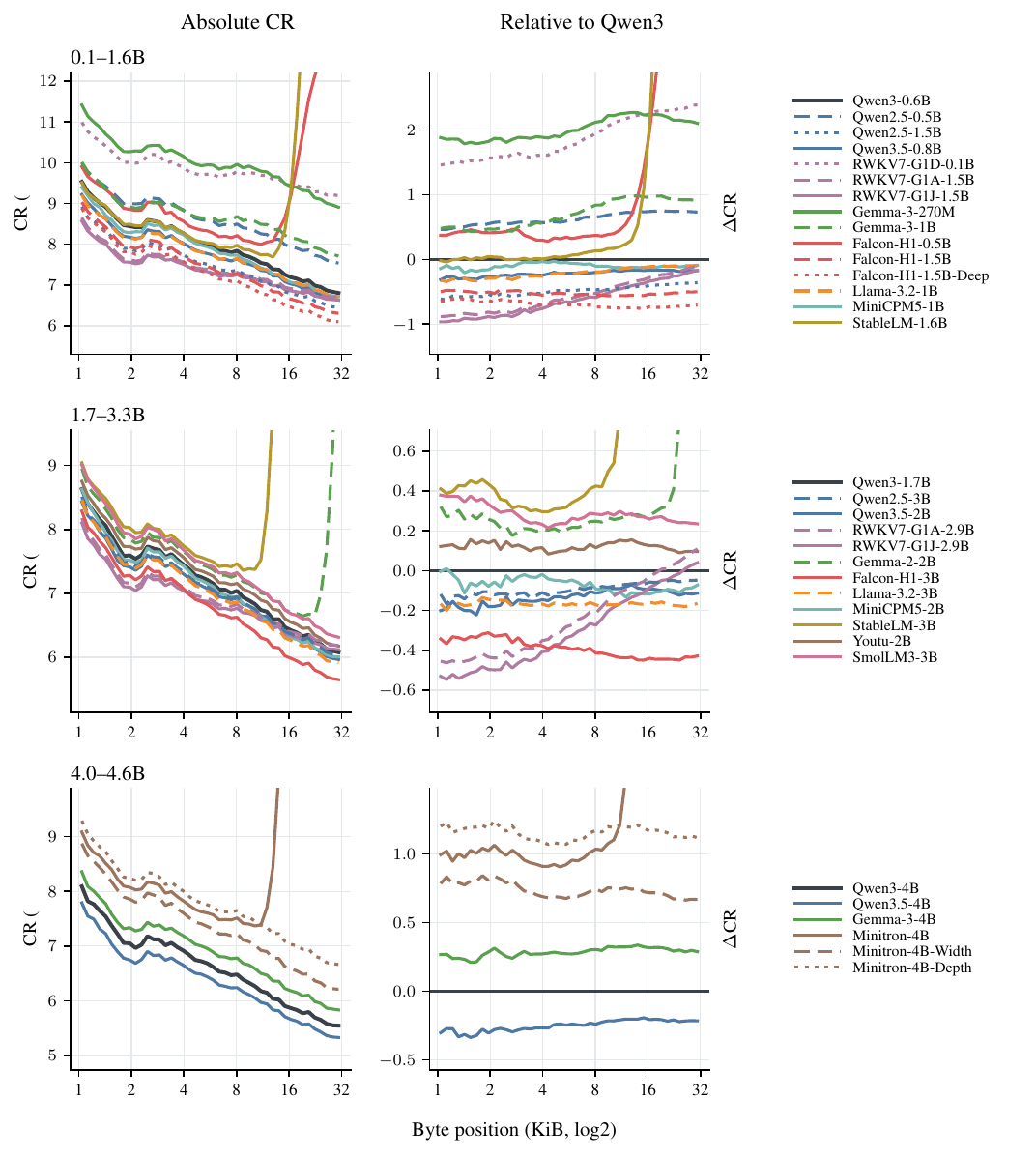}
\caption{Long-context CR for 33 models in the first three size groups. Each row shows absolute CR (left) and the difference from the Qwen3 model in that size group (right), pooled across four scientific-paper categories.}
\label{fig:long-context-scale}
\end{figure}
\clearpage
\begin{figure}[H]
\ContinuedFloat
\centering
\includegraphics[width=\linewidth]{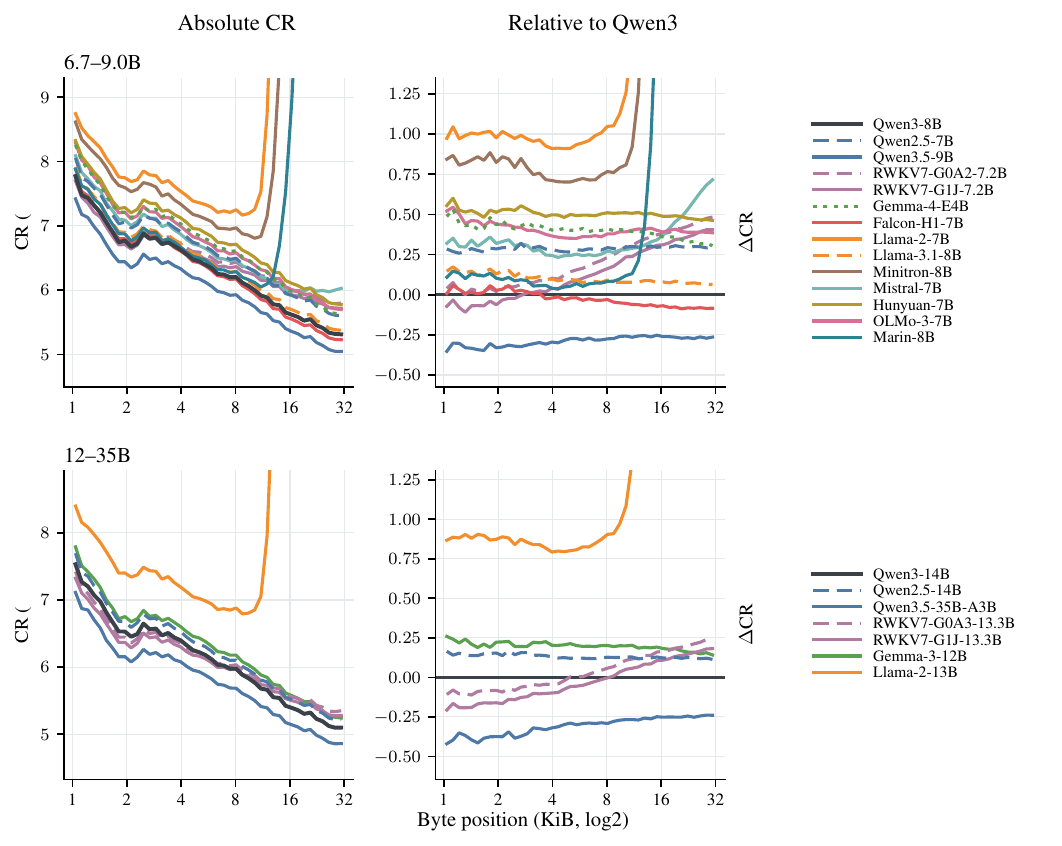}
\caption{Long-context CR for 21 models in the last two size groups. Each row shows absolute CR (left) and the difference from the Qwen3 model in that size group (right), pooled across four scientific-paper categories.}
\end{figure}
\clearpage
\section{Complete Category-Level Scaling Results}
\label{app:categories}

\begin{figure}[!htbp]
\centering
\includegraphics[width=\linewidth]{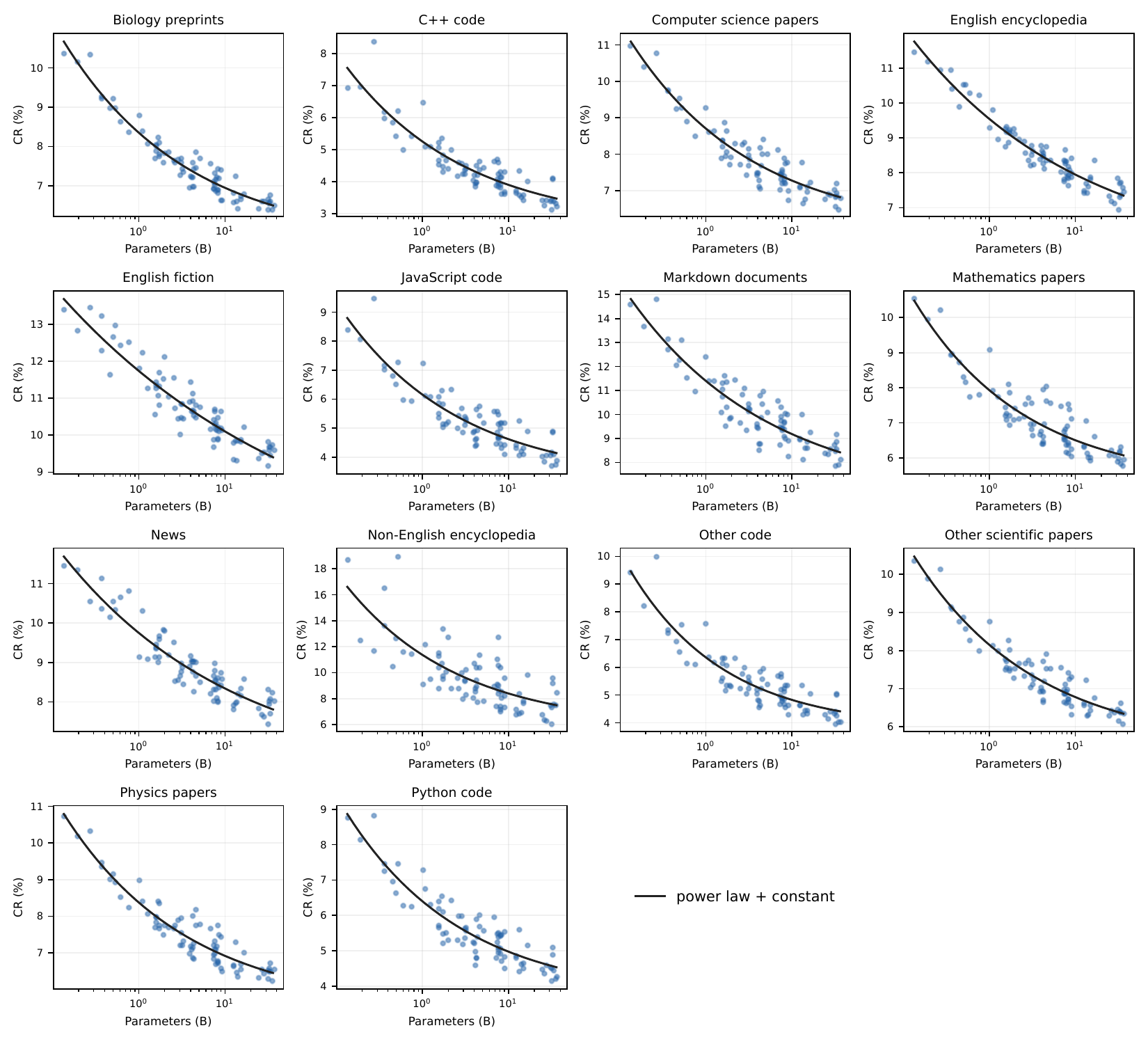}
\caption{Category-level scaling curves. Each panel shows CR for all 80 models and the fit defined in Eq.~\ref{eq:scaling-law}.}
\label{fig:category-curves}
\end{figure}
\FloatBarrier
\begin{table}[H]
\centering
\small
\caption{Category-level power-law-plus-constant fits to all 80 models in the July 2026 cohort.}
\label{tab:category-scaling}
\setlength{\tabcolsep}{5pt}
\begin{tabular}{lrrrrrr}
\toprule
Category & Median CR & $a$ & $b$ & $c$ & $R^2$ & RMSE \\
\midrule
Biology preprints & 7.315 & 2.801 & -0.301 & 5.546 & 0.935 & 0.229 \\
C++ code & 4.252 & 2.653 & -0.311 & 2.599 & 0.828 & 0.386 \\
Computer science papers & 7.720 & 2.827 & -0.305 & 5.878 & 0.884 & 0.321 \\
English encyclopedia & 8.380 & 4.057 & -0.217 & 5.487 & 0.928 & 0.266 \\
English fiction & 10.550 & 5.743 & -0.146 & 5.996 & 0.872 & 0.366 \\
JavaScript code & 5.122 & 2.972 & -0.317 & 3.184 & 0.808 & 0.469 \\
Markdown documents & 9.781 & 4.845 & -0.266 & 6.556 & 0.830 & 0.610 \\
Mathematics papers & 6.924 & 2.652 & -0.336 & 5.281 & 0.816 & 0.429 \\
News & 8.669 & 3.624 & -0.214 & 6.124 & 0.871 & 0.323 \\
Non-English encyclopedia & 9.155 & 5.602 & -0.329 & 5.768 & 0.594 & 1.549 \\
Other code & 5.249 & 2.565 & -0.395 & 3.792 & 0.819 & 0.473 \\
Other scientific papers & 7.165 & 2.693 & -0.309 & 5.459 & 0.894 & 0.294 \\
Physics papers & 7.293 & 2.906 & -0.302 & 5.465 & 0.891 & 0.316 \\
Python code & 5.397 & 2.731 & -0.321 & 3.670 & 0.833 & 0.400 \\
\bottomrule
\end{tabular}
\par\smallskip
\parbox{0.97\linewidth}{\footnotesize Parameters correspond to $\mathrm{CR}(P)=aP^b+c$, with $P$ measured in billions. RMSE is measured across all 80 models in CR percentage points.}
\end{table}

\section{Pareto-Frontier Scaling Results}
\label{app:pareto-scaling}

For the all-category score and each text type, we fit Eq.~\ref{eq:scaling-law} only to models on the parameter--CR Pareto frontier. A model lies on the frontier unless another evaluated model has at most as many parameters and lower CR.

\begin{figure}[!htbp]
\centering
\includegraphics[width=\linewidth]{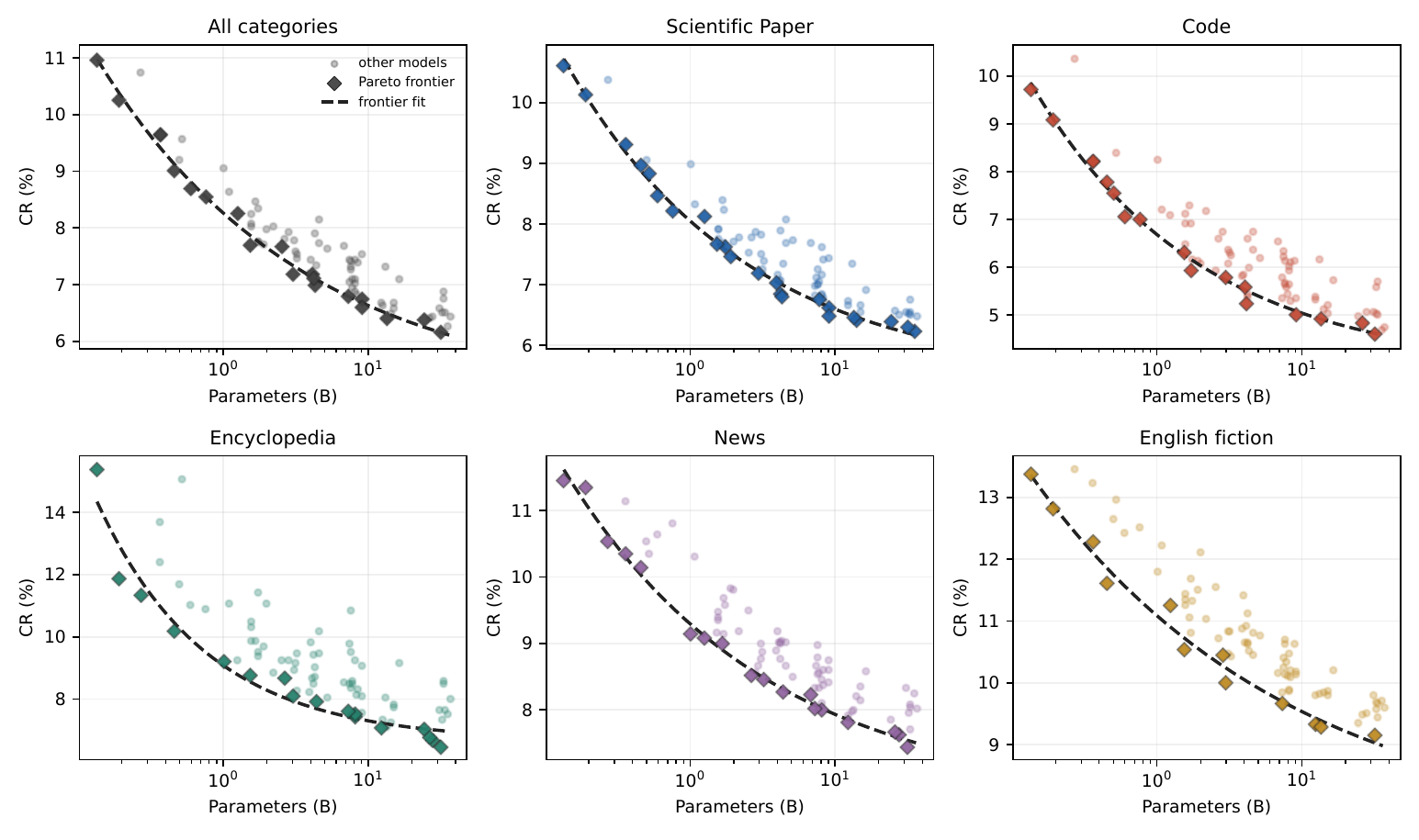}
\caption{Pareto-frontier scaling by text type. Diamonds mark frontier models, circles mark other models, and dashed lines fit the frontier models.}
\label{fig:pareto-scaling}
\end{figure}

\begin{table}[H]
\centering
\small
\caption{Pareto-frontier scaling fits by data type on the July 2026 80-model cohort.}
\label{tab:pareto-group-scaling}
\setlength{\tabcolsep}{5pt}
\begin{tabular}{lrrrrrrr}
\toprule
Type & Cats. & Frontier & $a$ & $b$ & $c$ & $R^2$ & RMSE \\
\midrule
All categories & 14 & 20 & 3.244 & -0.304 & 5.016 & 0.992 & 0.125 \\
Scientific Paper & 5 & 23 & 2.638 & -0.349 & 5.410 & 0.992 & 0.110 \\
Code & 5 & 17 & 2.853 & -0.372 & 3.831 & 0.991 & 0.148 \\
Encyclopedia & 2 & 17 & 2.421 & -0.575 & 6.661 & 0.960 & 0.454 \\
News & 1 & 18 & 2.657 & -0.313 & 6.636 & 0.993 & 0.105 \\
English fiction & 1 & 12 & 3.611 & -0.244 & 7.475 & 0.984 & 0.176 \\
\bottomrule
\end{tabular}
\end{table}

The all-category frontier contains 20 models from 0.135B to 31.3B parameters. Fitting these 20 models gives $\widehat{\mathrm{CR}}(P)=3.244P^{-0.304}+5.016$ ($R^2=0.992$; RMSE: 0.125 CR percentage points). Type-level frontiers contain 12--23 models, with exponents from $-0.575$ for encyclopedia articles to $-0.244$ for English fiction.
\FloatBarrier
\Needspace{24\baselineskip}
\section{Tokenization Statistics}
\label{app:tokenizer}

\begin{table}[H]
\centering
\small
\caption{Native-token segmentation diagnostics for the 80-model analysis cohort. Residual $\rho$ is the Spearman correlation between the residuals of CR and bytes per token after each is regressed on $\log_{10}$ parameter count with an intercept.}
\label{tab:tokenizer}
\setlength{\tabcolsep}{5pt}
\begin{tabular}{lrrrr}
\toprule
Category & Median bytes/token & Minimum & Maximum & Residual $\rho$ \\
\midrule
Biology preprints & 4.506 & 3.705 & 4.808 & 0.025 \\
C++ code & 3.357 & 2.719 & 3.731 & -0.194 \\
Computer science papers & 4.775 & 4.058 & 4.999 & 0.002 \\
English encyclopedia & 4.173 & 3.586 & 4.458 & -0.244 \\
English fiction & 4.305 & 3.755 & 4.498 & 0.114 \\
JavaScript code & 3.425 & 2.725 & 3.773 & -0.135 \\
Markdown documents & 3.607 & 2.884 & 3.871 & -0.207 \\
Mathematics papers & 3.576 & 3.144 & 3.864 & 0.228 \\
News & 4.601 & 3.991 & 4.830 & 0.100 \\
Non-English encyclopedia & 3.375 & 1.878 & 4.166 & -0.527 \\
Other code & 3.335 & 2.752 & 3.681 & -0.175 \\
Other scientific papers & 4.624 & 3.991 & 4.899 & 0.080 \\
Physics papers & 4.259 & 3.707 & 4.555 & 0.063 \\
Python code & 3.686 & 2.964 & 4.042 & -0.245 \\
\bottomrule
\end{tabular}
\end{table}

\FloatBarrier

\end{document}